\documentclass[journal]{IEEEtran}
\usepackage{ulem} 

\newcommand{\etal}{\textit{et al}.}
\newcommand{\ie}{\textit{i}.\textit{e}.}

\usepackage[misc]{ifsym}
\usepackage{algorithmicx,algorithm}
\usepackage{multirow}
\usepackage[noend]{algpseudocode}

\usepackage[pagebackref=true,breaklinks=true,linkcolor=red,anchorcolor=blue, citecolor=green,letterpaper=true,colorlinks,bookmarks=false]{hyperref}

\usepackage{breakurl}
\usepackage{graphicx}
\usepackage{amssymb}
\usepackage{bm}
\usepackage{bbm}
\usepackage{amsmath}
\usepackage{booktabs}
\usepackage{multirow}
\usepackage{makecell}
\usepackage{pifont}
\usepackage{bbding}

\usepackage{wrapfig}
\usepackage{verbatim}
\usepackage{relsize}

\usepackage{ulem}
\ifCLASSINFOpdf
\else
\fi

\begin{document}
\title{Rethinking Few-Shot Class-Incremental Learning with Open-Set Hypothesis in Hyperbolic Geometry}

\author{
Yawen Cui, Zitong Yu, Wei Peng, and Li Liu
\IEEEcompsocitemizethanks{\IEEEcompsocthanksitem This work was partially supported by National Key Research and Development Program of China No. 2021YFB3100800, the Academy of Finland under grant 331883 
and the National Natural Science Foundation of China under Grant 61872379, 62022091, and the China Scholarship Council (CSC) under grant 201903170129.
\IEEEcompsocthanksitem Li Liu is with the College of System Engineering, National University of Defense Technology (NUDT), Changsha, Hunan, China. She is also with the Center for Machine Vision and Signal analysis (CMVS), University of Oulu, Oulu, Finland.  Li Liu is the corresponding author.
(email:dreamliu2010@gmail.com)
\IEEEcompsocthanksitem Yawen Cui is with CMVS, University of Oulu, Oulu, Finland. (email: yawen.cui@oulu.fi)
\IEEEcompsocthanksitem Zitong Yu is with ROSE Lab, Nanyang Technological University, Singapore. (email: zitong.yu@oulu.fi)
\IEEEcompsocthanksitem Wei Peng is with CNSlab, Stanford University. (email: wepeng@stanford.edu)
}
}

% The paper headers
\markboth{In preparation for submitting to IEEE Transactions on Cybernetics}%
{Shell \MakeLowercase{\textit{et al.}}: Few-Shot Class-Incremental Learning}

% make the title area
\maketitle

\begin{abstract}
Few-Shot Class-Incremental Learning (FSCIL) aims at incrementally learning novel classes from a few labeled samples by avoiding the overfitting and catastrophic forgetting simultaneously. The current protocol of FSCIL is built by mimicking the general class-incremental learning setting, while it is not totally appropriate due to the different data configuration, \ie, novel classes are all in the limited data regime. In this paper, we rethink the configuration of FSCIL with the open-set hypothesis by reserving the possibility in the first session for incoming categories. To assign better performances on both close-set and open-set recognition to the model, Hyperbolic Reciprocal Point Learning module (Hyper-RPL) is built on Reciprocal Point Learning (RPL) with hyperbolic neural networks. Besides, for learning novel categories from limited labeled data, we incorporate a hyperbolic metric learning (Hyper-Metric) module into the distillation-based framework to alleviate the overfitting issue and better handle the trade-off issue between the preservation of old knowledge and the acquisition of new knowledge. The comprehensive assessments of the proposed configuration and modules on three benchmark datasets are executed to validate the effectiveness concerning three evaluation indicators.
\end{abstract}

\begin{IEEEkeywords}
Few-shot learning, class-incremental learning, hyperbolic deep neural network, open-set recognition, knowledge distillation
\end{IEEEkeywords}

\IEEEpeerreviewmaketitle

\section{Introduction}

Humans can generalize what they have learned to new encountered tasks without forgetting ancient knowledge dramatically, which is termed the capability of continual learning \cite{zenke2017continual,lopez2017gradient}. Simultaneously, they can also acquire the novel competence when provided limited information, \ie, the Few-Shot Learning (FSL) \cite{snell2017prototypical, finn2017model, cui2022coarse}. However, existing deep models suffer from severe catastrophic forgetting of previous tasks when adapting to novel ones. Furthermore, during the adaptation process, a huge number of labeled samples are required to guide the model training, which is computationally complex and impracticable for specific tasks. Inspired by humans' acquisition mode of knowledge or skills through experience, Few-Shot Continual Learning (FSCL) \cite{tao2020few} is put forward to narrow the gap between deep models and humans' cognitive system.

Continual learning~\cite{van2019three} is mainly implemented by three aspects of the increment: (1) Task-incremental learning aims at incrementally learning a sequence of disjoint tasks, which requires the task identity in the prediction procedure. (2) Class-incremental learning (CIL) is to construct a unified classifier for all encountered classes at different stages. (3) Domain-incremental learning targets to progressively learn categories in novel domains. In this paper, we focus on the class-incremental learning setting , \ie, Few-Shot Class-Incremental Learning (FSCIL). Compared with the other two configurations, FSCIL is the more practical, but also more challenging since the classifier is continually expanded with novel categories arriving. 

\begin{figure}[t]
  \centering
  %\fbox{\rule{0pt}{2in} \rule{0.9\linewidth}{0pt}}
   \includegraphics[width=1.0\linewidth]{ 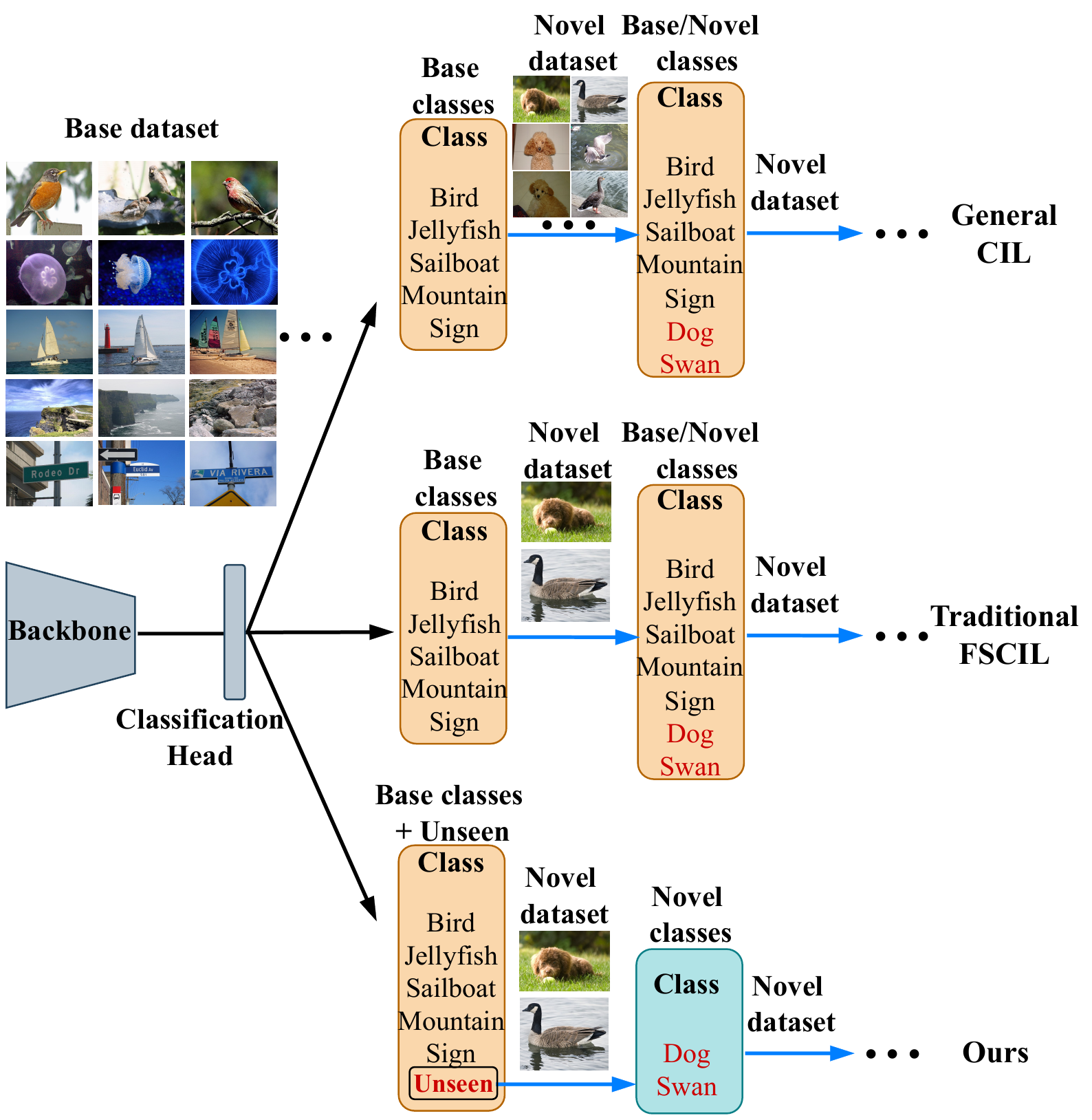}
   \vspace{-0.25in}
   \caption{The difference among general CIL, traditional FSCIL and Ours. We use the 2-way incremental paradigm as an example, \ie, two novel classes arrive in each incremental session. \textbf{General CIL}: the size of the dataset encountered in each session is totally the same or similar. \textbf{Traditional FSCIL}: The protocol of Traditional FSCIL imitates general CIL by treating the base and novel classes fairly in a unified classifier, while it is not totally suitable due to the different data configurations. For FSCIL, novel classes arriving in incremental sessions only contain limited labeled samples and we use 2-way 1-shot as an example here. \textbf{Ours}: We rethink the configuration of FSCIL with the open-set assumption. Specifically, we reserve the possibility in the first session for the incoming novel categories by regarding the these categories as the unseen ones. 
   }
   %\caption{Example of caption.
   %It is set in Roman so that mathematics (always set in Roman: $B \sin A = A \sin B$) may be included without an ugly clash.}
   \vspace{-0.2in}
   \label{fig:motivation}
\end{figure}

Given a model well-trained with a large-scale base dataset, FSCIL aims at incrementally learning novel classes from a few labeled samples by avoiding the overfitting and catastrophic forgetting simultaneously. The current protocol of FSCIL~\cite{tao2020few} is built by mimicking the general CIL setting~\cite{hou2018lifelong, rebuffi2017icarl}, while it is not totally appropriate due to the different data configurations.
For General CIL, the data of novel classes arriving in each incremental learning session is of the same quantity as the data of base classes in the first session. However, each incremental task in FSCIL is in the limited data regime, which exists severe data imbalance with the base session. Current FSCIL methods~\cite{tao2020few, zhang2021few, zhu2021self} treat base and novel classes in a fair manner, which results that the overall performance is dominated by the base classes and the performance disparity between base and novel classes is relatively large. 

Toward the aforementioned issues, we rethink the configuration of FSCIL with the open-set assumption, which is illustrated in Figure \ref{fig:motivation}. Specifically, due to the novel categories are encountered in the following sessions, we reserve the possibility in the first session for these incoming categories, which constitutes the open-set recognition problem \cite{scheirer2012toward, geng2020recent, chen2020learning}. Though this configuration is not consistent with the traditional CIL, it is more in line with real-world applications. We use the disease diagnosis as an example. For a particular case, the diagnosis system tends to rule out the underlying diseases first, then the rare diseases. When setting up the diagnosis system, we also need to leave the possibility for newly-emerging diseases to promote the evolution of the system. 

With the open-set assumption, in the base session (also termed the first session), the model need to be trained not for the discriminative ability of base categories, but for the open-set recognition by considering the novel categories arriving in the following incremental sessions as the open-world unknown categories. In open-set recognition tasks, the trade-off between close-set and open-set recognition always depends on the chosen thresholds, which is troublesome to handle this issue. However, for the FSCIL task, the model should be assigned better performances on both close-set and open-set recognition since we compute the overall classification accuracy on both base and novel categories. In this way, we put forward the Hyperbolic Reciprocal Point Learning module (Hyper-RPL) by optimizing Reciprocal Point Learning (RPL) \cite{chen2020learning} with hyperbolic geometry \cite{peng2021hyperbolic}. According to the distance of training samples and reciprocal points, RPL already provides a advanced optimization strategy with the classification loss on known categories and by reducing open-space risk. Particularly, in Hyper-RPL, we provide a richer representation space to search for the optima by integrating Euclidean space and hyperbolic space with the well-designed strategy. Due to the complementary representation in hyperbolic geometry, the open-set recognition capability is enhanced without undermining the performance of close-set recognition.

For the incremental learning sessions in our proposed configuration, we start with a new classification branch and exploit the distillation-based CIL framework. Compared with traditional FSCIL, there are mainly two merits: (1) With the previous FSCIL protocol and distillation-based CIL framework, Tao~\etal~\cite{tao2020few} point that the trade-off between old and new classes is more prominent due to the larger learning rate and stronger gradients when learning novel classes from limited labeled data. In contrast, by considering the open-set assumption in FSCIL, a new branch is designed for novel categories and the performance of the base categories can be completely preserved without updating the model parameters. (2) Distillation-based CIL framework requires to store samples of previously-seen categories for the replay in the following sessions. Since we do not need to store the samples of base classes concerning the catastrophic forgetting issue, the size of this extra memory in our proposed FSCIL configuration is smaller than that of the traditional FSCIL setting. 

Besides, to learn novel categories from limited labeled data, we incorporate a hyperbolic metric learning (Hyper-Metric) module into the distillation-based CIL framework. Compared with metric learning in the Euclidean space, Hyper-Metric module can learn the discriminative latent representation with intra-class compactness and inter-class sparseness, which can alleviate the overfitting issue and better handle the trade-off issue between the preservation of old knowledge and the acquisition of new knowledge. 

To sum up, the main contributions of this paper are listed:

\begin{itemize}
\item We rethink the FSCIL task with the open-set assumption, which is a more practical configuration.

\item The Hyper-RPL module is proposed to enhance the ability of open-set recognition without undermining the performance of close-set recognition.

\item To alleviate the overfitting issue, Hyper-Metric is proposed by incorporating the hyperbolic metric learning into distillation-based CIL framework.  

\item We provide a comprehensive assessment of the proposed configuration and modules on three benchmark datasets to validate the effectiveness concerning three evaluation indicators.
\end{itemize}

\section{Related Work}

\subsection{Few-Shot Learning}
The aim of Few-Shot Learning (FSL) is to tackle the target task with limited labeled samples per class. A hypothesis for FSL is that the target task can access a source/base dataset with large-scale labeled training instances, and there is no class-level or data-level intersection between the source dataset and the target dataset. Traditional FSL can be categorized into data augmentation-based methods, meta-learning-based methods, and metric learning-based methods.
Data augmentation-based methods~\cite{schwartz2018delta} aim to synthesize more data from the novel classes to facilitate the regular learning. Schwartz \etal use a variant of auto-encoder to capture the intra-class difference between two class samples in the latent space, the transfer class distributions from training to novel classes. The mechanism of metric learning-based methods is to learn a semantic embedding space using a distance loss function. Matching networks~\cite{vinyals2016matching} maximizes the cross-entropy for the non-parametric softmax classifier. In prototypical networks~\cite{snell2017prototypical}, each category is represented by the prototype, and the objective is to maximize the cross-entropy with the prototypes-based probability expression. Meta-learning-based methods target to learn a learning strategy to adjust well to a new FSL task. MAML \cite{finn2017model} aims at obtaining optimal initialization parameters of the model for a novel task through meta-training.

\subsection{Class-Incremental Learning}
Class-Incremental Learning (CIL) is the most challenging scenario of continual learning~\cite{van2019three}, which targets to learning a classifier incrementally to recognize all encountered categories. To solve the CIL tasks, the proposed methods mainly follow the two strategies. The first one is to identify and preserve significant parameters of the previous model~\cite{kirkpatrick2017overcoming, kaushik2021understanding}. Kirkpatrick \etal \cite{kirkpatrick2017overcoming} propose to remember old tasks by selectively learning the weights that are important for those tasks. Kaushik~\etal~\cite{kaushik2021understanding} introduce Relevance Mapping Networks (RMNs), which reflects the relevance of the weights
for the current task by assigning large weights to
essential parameters. The second strategy is preserving the knowledge of the old model through some strategies like knowledge distillation \cite{rebuffi2017icarl, hou2019learning}. iCaRL~\cite{rebuffi2017icarl} learns strong classifiers and a data representation simultaneously. Hou~\etal~\cite{hou2019learning} develop
a new framework for incrementally learning a unified classifier which treats both old and new classes
uniformly.

Tao~\etal~\cite{tao2020few} first introduce few-shot learning into CIL, which is formulated as Few-Shot Class-Incremental Learning (FSCIL)~\cite{zhang2021few,chen2021incremental,cheraghian2021semantic}. It is a more practical configuration that aims at incrementally learning novel classes with limited labeled samples. Existing solutions can be categorized into two types: one refers to knowledge representation and refinement, and another one is knowledge distillation-based. As for the first type, Chen~\etal~\cite{chen2021incremental} propose a nonparametric method where knowledge about the learned tasks is compressed into a
small number of quantized reference vectors. CEC~\cite{zhang2021few} is proposed by employing a graph model to propagate context information between classifiers for adaptation. Relying on knowledge distillation technique, Cheraghian \etal \cite{cheraghian2021semantic} employs the word embedding as the semantic information to
facilitate the distillation process. Considering the trade-off between old-knowledge preserving and new-knowledge adaptation in FSCIL, Dong~\etal~\cite{dong2021few} propose the exemplar relation distillation incremental learning framework.

\subsection{Hyperbolic Geometry}

As a counterpart in the Euclidean space, hyperbolic geometry is no-Euclidean geometry with a constant negative Gaussian curvature. Khrulkov \cite{khrulkov2020hyperbolic} pointed out that hyperbolic geometry can be beneficial in analyzing image manifolds by capturing both semantic similarities and hierarchical relationships between images. The hyperbolic geometry has been utilized for representing tree-like structures and taxonomies~\cite{ganea2018hyperbolic, nickel2017poincare, nickel2018learning}, text~\cite{aly2019every, tifrea2018poincare}, and
graphs~\cite{bachmann2020constant, bachmann2020constant, liu2019hyperbolic}. There are five isometric models of hyperbolic geometry: the Hyperboloid model, the Klein model, the Hemisphere model, the Poincaré ball model, and the Poincaré half-space model. In this paper, we appply the Poincaré ball to model the hyperbolic embedding space. 
Hyperbolic deep neural networks is the neural structures built in the space of hyperbolic geometry, which can provide the powerful geometrical representations. Nickel and Kiela~\cite{nickel2017poincare} conducted the pioneering work on learning representation in hyperbolic spaces. Then, hyperbolic neural networks is first proposed by the work~\cite{ganea2018hyperbolic}, which introduce hyperbolic geometry into deep learning. Recent works of hyperbolic image embeddings \cite{khrulkov2020hyperbolic, liu2020hyperbolic} add one or two hyperbolic layers~\cite{ganea2018hyperbolic} after Euclidean deep convolutional neural networks. In this paper, we adopt the hyperbolic operation in the work~\cite{khrulkov2020hyperbolic} for the hyperbolic embedding.

\subsection{Open-Set Recognition}
The target of open-set recognition is to identify whether a given object instance belongs to a known class or not. This task was first introduced by the work~\cite{scheirer2012toward} where it was formalized as a constrained minimization problem based on Support Vector Machines (SVMs). The current works for open-set recognition mainly follow two lines: discriminative models and generative models. Discriminative models can be further
categorized into traditional machine learning-based methods and
deep learning-based methods. Based on SVM, the pioneering work~\cite{scheirer2012toward} propose a novel 1-vs-set machine to model a decision space. In the subsequent works, other more traditional approaches, such as Extreme Value Theory~\cite{jain2014multi}, sparse representation~\cite{zhang2016sparse}, and Nearest Neighbors~\cite{mendes2017nearest}, were utilized for open-set recognition. Owing to the strong representation ability, deep learning-based methods emerge with more attention in recent years. The first deep learning-based work~\cite{bendale2016towards} proposes an
Openmax function by calibrating the Softmax probability of each class with a Weibull distribution model. Yoshihashi~\etal~\cite{yoshihashi2019classification} adopt the Classification-Reconstruction training for Open-Set Recognition (CROSR), which utilizes latent representations for reconstruction and enables robust unknown detection and the known-class classification. 
However, this line of work faces the challenge of threshold tuning since a threshold is required to separate known and unknown classes. 

Other line of work employs generative models to synthesis the data of unseen classes or forecast the distribution of of novel classes. G-OpenMax~\cite{ge2017generative} extends Openmax by introducing a generative adversarial network for producing synthetic samples of novel categories. Class conditional auto-encoder (C2AE)~\cite{oza2019c2ae} considered the open-set recognition as a two-step problem, \ie, the encoder learns the closed-set classification and the decoder learns the open-set recognition task by reconstructing conditioned on class identity. \cite{perera2020generative} applied self-supervision and augmented the input
features obtained from a generative model to force class activations of open-set samples to be low. Zhou~\etal~\cite{zhou2021learning} proposed to prepare for the unknown classes by allocating placeholders for both data and classifier. 

\begin{figure}[t]
  \centering
  %\fbox{\rule{0pt}{2in} \rule{0.9\linewidth}{0pt}}
   \includegraphics[width=0.8\linewidth]{ 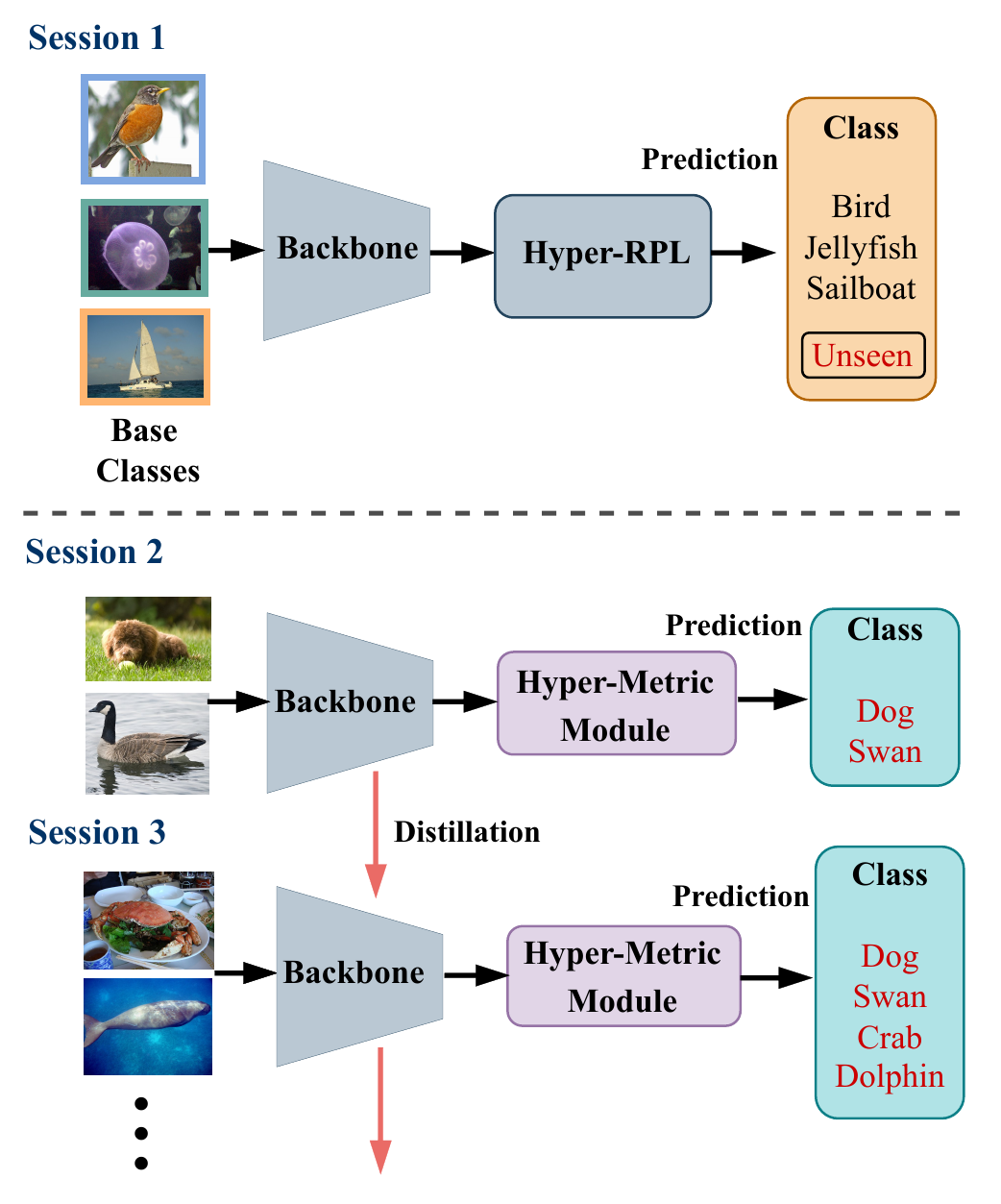}
   \vspace{-0.2in}
   \caption{ The proposed framework with Hyper-RPL module and Hyper-Metric module. For the first session, we reserve the possibility of novel classes by regarding them as the unseen categories. Hyper-RPL module is proposed for the close-set and open-set recognition task. For the following sessions, we build a new branch incrementally for novel categories. Hyper-Metric module is combined into the distillation-based framework to alleviate the overfitting issue with a expanded representation space. 
   }
   %\caption{Example of caption.
   %It is set in Roman so that mathematics (always set in Roman: $B \sin A = A \sin B$) may be included without an ugly clash.}
   \vspace{-0.15in}
   \label{fig:method}
\end{figure}

\section{Methodology}
In this section, we introduce the novel configuration of FSCIL with details first. Then, based on this, we propose an efficient solution for FSCIL. Our framework illustrated in Figure~\ref{fig:method} contains two key components: Hyperbolic Reciprocal Point Learning for open-set recognition in the base session, and Hyperbolic Metric Learning for alleviating the overfitting issue of novel classes. We first provide the problem formation with learning targets, then the overview of the framework, and finally the details of two components in the framework will be introduced.

\subsection{Problem Formulation}
Given a sequence of disjoint datasets by $\mathcal{D}=\{\mathcal{D}_1, \mathcal{D}_2, ...,\mathcal{D}_n\}$, here $\mathcal{D}_{1}$ is the large-scale base dataset used in the first base session and the followings are all novel few-shot datasets. 
Specifically, we term $\mathcal{D}_{1}=\left\{\left(\bm{x}_j, y_j\right)\right\}_{j=1}^{|\mathcal{D}_1|}$ where $ y_j \in \mathcal{C}_1$, and $\mathcal{C}_1$ represents the base category set. In the $i$-th session where $i>1$, the novel/new dataset is defined as $\mathcal{D}_{i}=\left\{\left(\bm{x}_j, y_j\right)\right\}^{N \times K}_{j=1}$ consists of $N$ classes $\mathcal{C}_i$ with $K$ labeled examples per class, \ie, a $N$-way $K$-shot problem. It is worth noting that there is no overlap between the categories and samples of different sessions, \ie, $\mathcal{C}_{i} \cap \mathcal{C}_{i'} = \emptyset$ and $\mathcal{D}_{i} \cap \mathcal{D}_{i'} = \emptyset$, where $i\neq i'$. In this paper, we arrange and verify our proposed configuration and solution on object classification task. 
%As for the model $\mathcal{F}\left(\cdot\right)$, it usually contains the backbone $\Theta\left(\cdot\right)$ for extracting features and the classification head $\Gamma\left(\cdot\right)$. 

The objective of FSCIL with our novel configuration is to train a open-set recognition model in the first session, then a sequence of disjoint novel datasets will be operated incrementally in a new branch. In this way, the discriminative ability of first session can be totally preserved and the performance imbalance caused by data imbalance can be mitigated. When performing the prediction, the sample is sent to the base branch first to check if it belongs to the base category or not. If not, it will be introduced into the novel branch.

\subsection{Overview of the Framework}
Based on open-set hypothesis, we rebuild the FSCIL configuration by considering the possibility of novel classes in the base session, and this novel framework together with our proposed solution is demonstrated in Figure~\ref{fig:method}. 

For the first session, we conduct a open-set recognition task with a $|C_1| + 1$ classifier where the extra one category stands for the unseen categories currently. To achieve this, we propose a hyperbolic reciprocal point learning (Hyper-RPL) by combining the hyperbolic geometry into the existing RPL algorithm for a more efficacious open-set recognition ability. When it comes to the $i$-th session where $i>1$, a new classification branch is created for the novel categories, which follows an incremental manner corresponding to the sequence of novel few-shot learning tasks. Due to the new proposed configuration, the trade-off issue between base categories and novel categories can be better alleviated. Because of this, advanced learning  strategies can be incorporated into the newly-created branch. In this way, we introduce the metric learning module with hyperbolic geometry into commonly-used cross-entropy loss. Moreover, in this branch, catastrophic forgetting issue is also encountered, and we utilize the knowledge distillation technique to tackle this issue. 

\subsection{Hyperbolic Reciprocal Point Learning}

For the first session, we conduct the open-set recognition to classify the base categories and regard the novel categories of the following sessions as the unseen class. Existing methods of open-set recognition suffer from the trade-off issue on the performance of close-set classes and open-set classes since a threshold is rendered to decide when a test sample is identified as the unseen categories. With guaranteed classification performance of close-set classes, we propose the Hyperbolic Reciprocal Point Learning (Hyper-RPL) module base on Reciprocal Point Learning (RPL) \cite{chen2020learning} to enhance the discriminative ability of unseen classes. 

\subsubsection{Preliminaries}
Here we give the description of RPL. Commonly, we use prototypes or class means to represent each class in the latent space. Conversely, Reciprocal Point is a novel concept that denote the extra-class space corresponding to each known category, which means that it stands for the reciprocal space of prototypes or class means. A sample can be recognized as known or unknown based on the distance with reciprocal points. We denote the latent representation space of category $k$ as $\mathcal{S}_k$ and its corresponding open-space as $\mathcal{O}_k = \mathbb{R}^d - \mathcal{S}_k$, where $\mathbb{R}^d$ is the d-dimensional full space. The open space $\mathcal{O}_k$ can be divided into positive open space $\mathcal{O}_k^{pos}$ which is from other known classes, and positive open space $\mathcal{O}_k^{neg}$ which stands for the remaining infinite unknown space, \ie, $\mathcal{O}_k=\mathcal{O}_k^{neg} \cup \mathcal{O}_k^{neg}$. With $|C_1|$ categories in the first session, we construct the entire learning process based on the open-set recognition of a single class. Then, the positive training data $\mathcal{D}_{1,k}$ contains samples from categories $k$, and samples from other known classes are regarded as negative training data $\mathcal{D}_{1, {\neq k}}$. Except $\mathcal{D}_1$, samples from $\mathbb{R}^d$ are the potential unknown data $\mathcal{D}^U$. The target of one-class open-set recognition task is to optimize a binary prediction function $\psi_k$ by minimizing the error $R_k$ defined as:
\begin{equation}
\mathop{\mathrm{arg} \space \mathrm{min}}_{\psi_k } \left \{ R_k | ( R_k^\epsilon\left ( \psi_k, \mathcal{S}_k \cup \mathcal{O}_k^{pos} \right ) + \alpha \cdot R_k^o\left (\psi_k, \mathcal{O}_k^{neg}  \right )  \right \},
\label{eq:one}
\end{equation}
where $\alpha$ is a positive regularization parameter. $R_k^\epsilon$ represents the empirical classification risk on known data, and $R_k^o$ is the open space risk function. Based on the Equation~\ref{eq:one}, the error ${\textstyle \sum_{|C_1|}^{k=1}} \mathit{R}_k$ of the entire learning process is formulated as 
\begin{equation}
\mathop{{\mathrm{arg} \space \mathrm{min}}}_{f} \left \{ \mathit{R}^\epsilon\left (f, \mathcal{D}_1  \right ) + \alpha \cdot  {\textstyle \sum_{|C_1|}^{k=1}} \mathcal{R}^o\left ( f, \mathcal{D}^U \right )     \right \}, 
\label{eq:all}
\end{equation}
where $f$ is the multi-class classification function. 

In order to represent the open space of category $k$, a set of reciprocal points for category $k$ is proposed and denoted as $\mathcal{P}_k=\left\{p _{k,i}|i=1,2,...,M \right \}$, where $M$ is the size of reciprocal point set. According to the distance of training samples and reciprocal points, the open-set network is optimized with targets: minimizing the classification loss on known categories and reducing open-space risk.

\subsubsection{Optimization with Hyperbolic Geometry}

\begin{figure}[t]
  \centering
  %\fbox{\rule{0pt}{2in} \rule{0.9\linewidth}{0pt}}
   \includegraphics[width=1.0\linewidth]{ 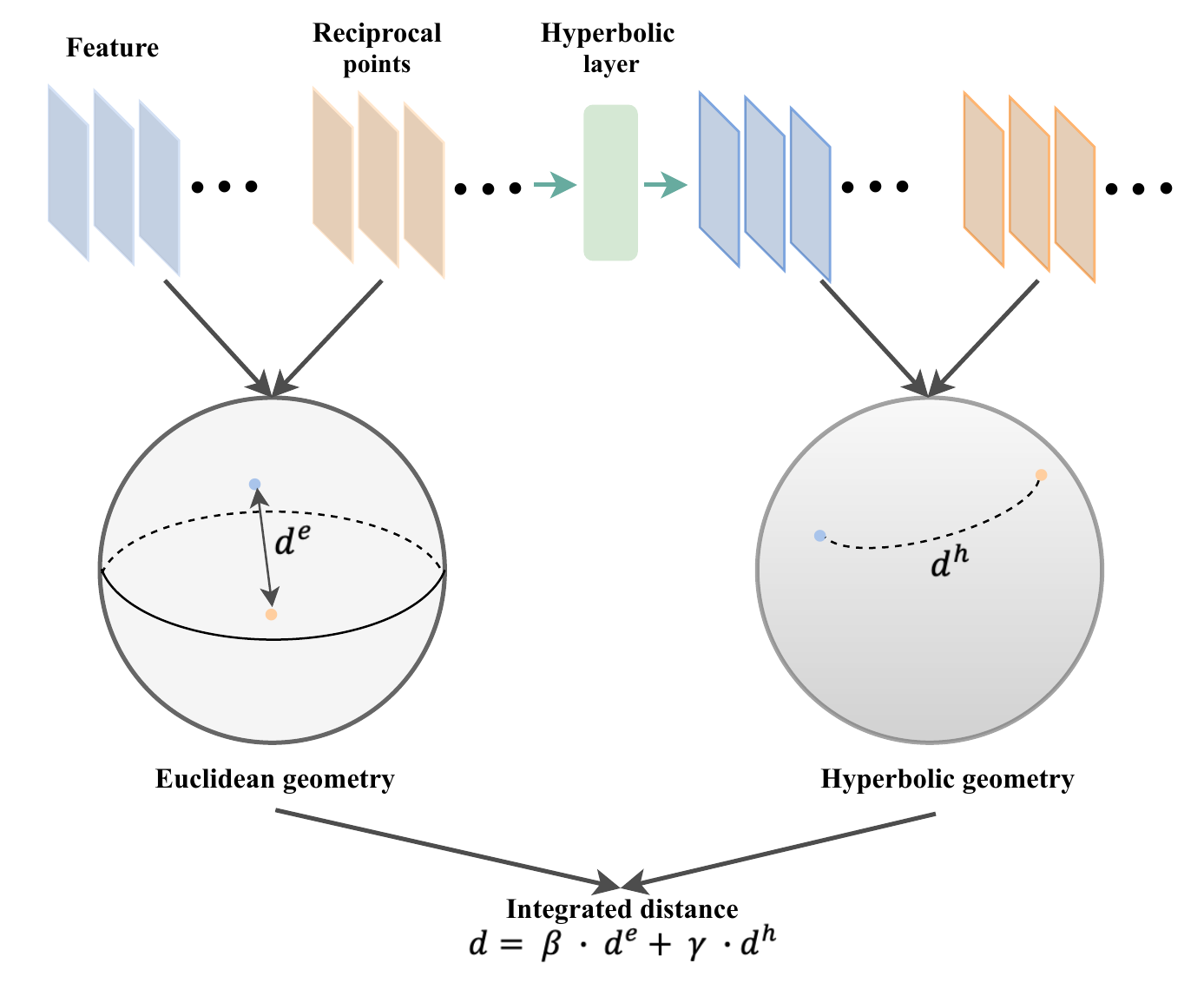}
   \vspace{-0.30in}
   \caption{The distance integration in Hyper-RPL module.
   }
   %\caption{Example of caption.
   %It is set in Roman so that mathematics (always set in Roman: $B \sin A = A \sin B$) may be included without an ugly clash.}
   \vspace{-0.2in}
   \label{fig:distance}
\end{figure}

For detecting unknown classes in RPL, it is pointed out that unknown samples have a closer distance with all reciprocal points than training samples of known classes. Further, for a specific test instance, the probability that it belongs to a known class is proportional to the distance between this instance and the farthest reciprocal point of this known class. In this way, a suitable operating threshold should be assigned to the model for identifying the known and unknown classes. However, RPL is very sensitive for this chosen threshold, \ie, the trade-off between the performance of known and unknown classes is intractable. To tackle this issue, we propose to optimize RPL by incorporating the hyperbolic geometry, which provides a richer search space for the optima that achieve a better separation of known and unknown space.

\begin{table*}[]
\caption{The dataset protocols for FSCIL. \#Categories and \#Samples stand for the number of categories and the number of samples per category, respectively. The learning pattern represents the setting of novel tasks in each incremental learning session.}
 \vspace{-0.1in}
\centering
\setlength{\tabcolsep}{11pt}
\resizebox{12.0cm}{!}{
\tabcolsep 0.06in
\begin{tabular}{ccc|ccc}
\bottomrule[1.3pt]
 \multirow{2}{*}{\textbf{Dataset}}   & \multicolumn{2}{c|}{\textbf{Base session}} & \multicolumn{3}{c}{\textbf{Incremental session}} \\ \cline{2-6} 
                        &    \#Categories       &   \#Samples  &   \#Categories    &   \#Samples    &   Incremental pattern   \\ \cline{1-6} 
 \textbf{CIFAR100}                             &     60      &     500      & 40      &    5   &   5-Way 5-Shot   \\ 
\textbf{\textit{mini}ImageNet}                             &    60       & 500      &   40    &    5   &  5-Way 5-Shot    \\ 
\textbf{CUB200}   &                   100       &       30    &     100     &   5    &   10-Way 5-Shot         \\                             
\bottomrule[1.3pt]
\end{tabular}
}
\vspace{-0.2in}
\label{table:dataset}
\end{table*}

For the classification loss of known classes, the target is to maximize the distance between the reciprocal points of the category and its corresponding training known samples. We denote the Euclidean embedding function of the open-set network as $f_\theta$ with trainable parameters $\theta$, and the transformation function to the hyperbolic geometry as $h_\varphi$ with parameters $\varphi$. In this paper, we employ Poincaré ball model to display the $d$-dimensional hyperbolic space. The Poincaré ball model $(\mathbb{D}^d_c, g^{{\mathbb{D}}_c})$ is defined by the manifold $\mathbb{D}^d_c =\{ \mathbf{x} \in \mathbb{R}^d:c\left \| \mathbf{x}  \right \|^2 <1, c\ge 0\}$ endowed with the Riemannian metric $g^\mathbb{D}(\mathbf{x})= \lambda^2_\mathbf{x}g^E$, where $||\cdot||$ denotes the Euclidean norm, and $c$ is the curvature of Poincaré ball model. $\lambda_\mathbf{x} =  \frac{2}{1-c||\mathbf{x}||^2}$ is the conformal factor, and $g^E$ is the Euclidean metric tensor. Formally, the geodesic distance between two points in the Poincaré ball is defined as 
\begin{equation}
d^p(\mathbf{x},\mathbf{y})=\frac{2}{\sqrt{c}}\mathrm{arctanh}(\sqrt{c}\left \| -\mathbf{x}\oplus_c \mathbf{y}  \right \| ),  
\label{eq:h_d}
\end{equation}
where $\oplus$ is the M\"obius addition. 

As shown in Figure~\ref{fig:distance}, given a training sample $x$ of category $k$ in $\mathcal{D}_1^k$, the integrated distance between it and reciprocal points $\mathcal{P}_k$ is formulated as 
\begin{equation}
\begin{split}
d\left(f_\theta \left ( x \right ), h_\varphi,  \mathcal{P}_k \right) =  \beta \cdot  d^e\left (f_\theta \left ( x \right ) , \mathcal{P}_k \right ) \\
+ \gamma \cdot d^h\left ( h_\varphi,  f_\theta \left ( x \right ), \mathcal{P}_k  \right ),
\label{eq:d_all}
\end{split}
\end{equation}
where $d^e(f_\theta (x), h_\varphi,  \mathcal{P}_k)$ is the distance function in Euclidean geometry,  $d^h(h_\varphi,  f_\theta (x), \mathcal{P}_k)$ is the distance function in hyperbolic geometry, and $\beta$ and $\gamma$ are the integrated weights. In addition, $\beta + \gamma = 1$. $d^e(f_\theta (x), h_\varphi,  \mathcal{P}_k)$ is formulated as 
\begin{equation}
d^e\left ( f_\theta \left ( x \right ), \mathcal{P}_k  \right) = \frac{1}{M} {\textstyle \sum_{i=1}^{M}} \left \| f_\theta\left ( x \right ) - \mathcal{P}_{k,i}  \right \|_2^2.
\label{eq:d_e}
\end{equation}

Here the $h_\varphi$ is assigned by $g^{\mathbb{D}}$. For the distance in hyperbolic space, the first procedure is to transform $f_\theta(x)$ and $\mathcal{P}_k$ into hyperbolic space. Then, $d^h( h_\varphi,  f_\theta \left ( x \right ), \mathcal{P}_k)$ is defined as 
\begin{equation}
d^h( h_\varphi,  f_\theta \left ( x \right ), \mathcal{P}_k) = \frac{1}{M}{\textstyle \sum_{1}^{M}} d^p(h_\varphi (f(x), h_\varphi (\mathcal{P}_{k,i}))).
\label{eq:d_h}
\end{equation}

With Equation~\ref{eq:d_all}, the probability $p\left (y=k|x, f_\theta, \mathcal{P}_k, h_\varphi  \right ) $ that $x$ belongs to category $k$ is computed based on the softmax function. After that, the classification loss of known classes is given by
\begin{equation}
\mathcal{L}^c_k(x, \theta,\varphi)=-log\; p\left (y=k|x, f_\theta, \mathcal{P}_k, h_\varphi  \right ).
\label{eq:loss_c}
\end{equation}

For the target of reducing open-space risk, we first define a global multiple class-wise open spaces $\mathcal{O}^G$ as the intersection of open spaces of all known categories. RPL proposes to restrict the open space to a bounded range, namely, the distance between samples from $\mathcal{S}_k$ and $\mathcal{P}_k$ is constrained to a certain extent. In this paper, we formulate the open space risk as
\begin{equation}
\mathcal{L}^o_k(x, \theta,\varphi, \mathcal{R}_k) = \frac{1}{M}{\textstyle \sum_{1}^{M}}||d\left(f_\theta \left ( x \right ), h_\varphi, \mathcal{P}_k \right)-\mathcal{R}_k||^2_2,
\label{eq:os-loss}
\end{equation}
where $\mathcal{R}_k$ is the learnable margin. Finally, the total loss of the base session is defined as
\begin{equation}
\mathcal{L}^B = \sum^{|C_1|}_{k=1}(\mathcal{L}_k^c + \lambda\mathcal{L}^o_k),
\label{eq:all-loss}
\end{equation}
where $\lambda$ is a hyper-parameter that determines the weight of reducing the open-space risk. 

With Equation~\ref{eq:all-loss} and the integrated distance function, we provide a more comprehensive space to search for the optima by reducing the classification loss of known classes and open-space risk. In this way,  without deteriorating the performance of close-set classification performance, the detection performance of unseen classes can be enhanced.

\subsection{Hyperbolic Metric Learning}
For FSCIL, Tao~\etal point out the class-imbalance issue and the performance trade-off issue between old and new classes. For the first issue,  the output logits are biased towards
base classes with large-scale training datasets. The trade-off issue mainly results from a larger learning rate for learning from very few training samples, which makes maintaining the output for old classes very difficult.  
Since a new branch is set for incremental learning sessions in the proposed framework, we do not need to consider preserving the performance of base categories and thus more advanced learning strategies can be combined into the distillation-based framework for learning novel classes. Nevertheless, novel classes are in the limited data regime and thus easily prone to the overfitting issue. If we incorporated more learning strategies, overfitting issue may become more severe. In this paper, we introduce Hyperbolic Metric learning (Hyper-Metric) module to provide a more comprehensive learning space for novel classes to mitigate the overfitting issue. Hyper-Metric is to conduct the metric learning in hyperbolic geometry, which targets at pushing the representations of the same class to be closer and others to be farther. Specifically, we apply the pair-wise cross-entropy loss to achieve this, and then incorporate this metric learning loss into the distillation-based class-incremental learning framework.  

For each epoch, we sample $M$ positive pairs, and each positive pair $(x_i,x_j)$ contains two samples from the same category. The total number of samples is $T=2M$. Here the loss function for a specific positive pair is defined as 
\begin{equation}
\mathcal{L}^m_{i,j} = -log\frac{exp(d^p(h_\varphi(f(x_i)), h_\varphi(f(x_j)))/\tau )}{ {\textstyle \sum_{t=1,t\ne i}^{T}} exp(-d^p(h_\varphi(f(x_i)), h_\varphi(f(x_t)))/\tau )},  
\label{eq:metric_loss}
\end{equation}
where $\tau$ is the temperature hyper-parameter. $\tau$ affects the smoothness degree of obtained logits, and values are more smooth when $\tau$ is larger.
For a batch, we obtain the total metric loss $\mathcal{L}^m$ on all positions pairs including  $(x_i,x_j)$ and $(x_j,x_i)$. When combined the Hyper-Metric module into the distillation-based incremental learning framework, we define the total loss of incremental learning sessions as
\begin{equation}
\mathcal{L}^I = \mathcal{L}^{ce} + \zeta\mathcal{L}^{dl} + \eta \mathcal{L}^m,
\label{eq:incremental-loss}
\end{equation}
where $\mathcal{L}^{ce}$ is the cross-entropy loss, $\mathcal{L}^{dl}$ is the general distillation loss, and $\eta$ and $\zeta$ are the loss weights. Moreover, $\mathcal{L}^{dl}$ and $\mathcal{L}^{ce}$ are implemented based on iCaRL~\cite{rebuffi2017icarl}.

By introducing the Hyper-Metric module into the distillation-based class-incremental learning framework, the classification performance can be improved to some extent. Compared with the performance obtained with metric learning in Euclidean space, the extra hyperbolic learning space can alleviate the overfitting issue.  

%\subsection{Inference Procedure}

\begin{figure*}[t]
  \centering
  %\fbox{\rule{0pt}{2in} \rule{0.9\linewidth}{0pt}}
   \includegraphics[width=0.75\linewidth]{ 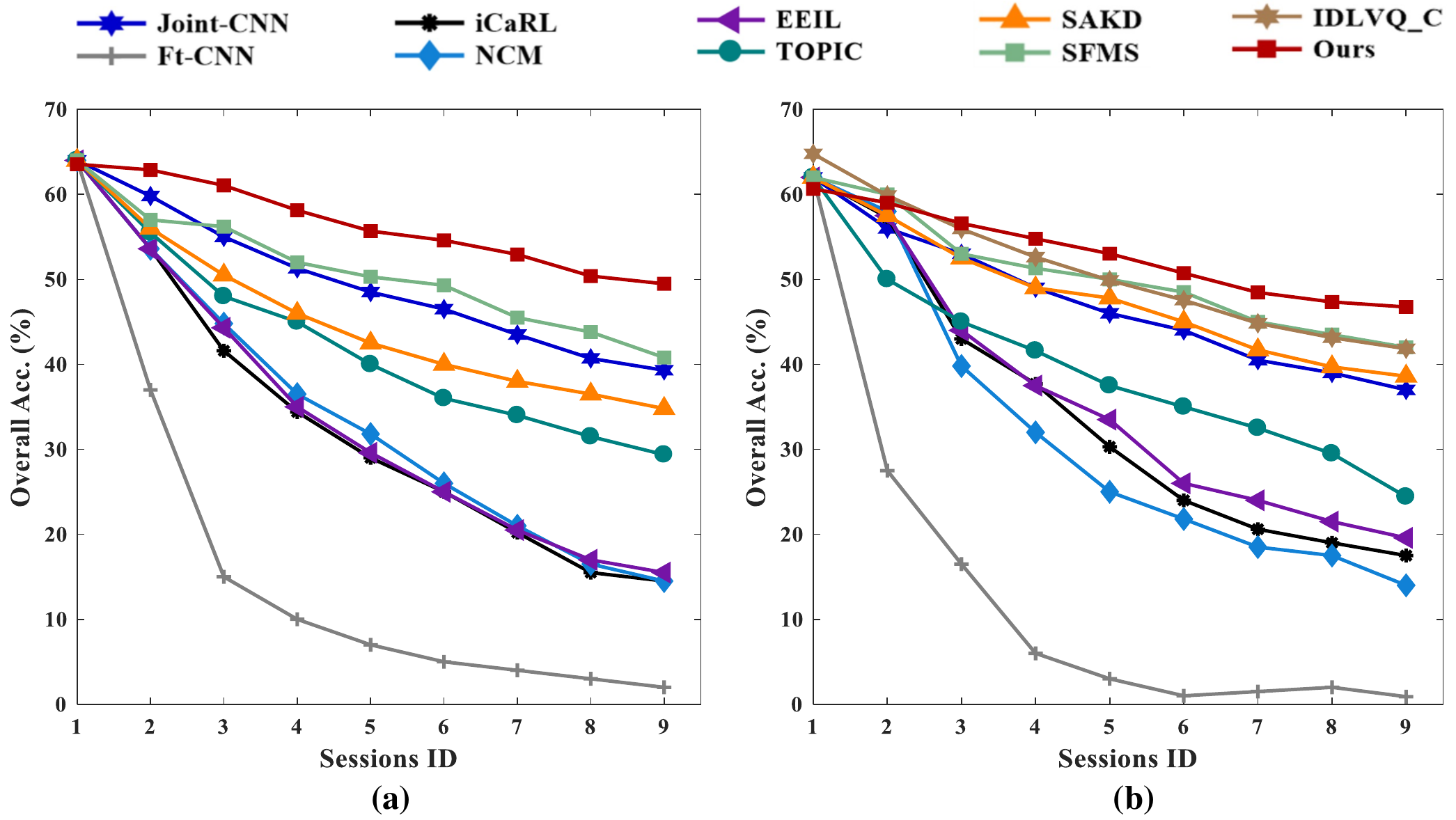}
   \vspace{-0.1in}
   \caption{The comparative results of Joint CNN~\cite{tao2020few}, Ft-CNN~\cite{tao2020few}, iCaRL~\cite{rebuffi2017icarl}, NCM~\cite{hou2019learning}, EEIL~\cite{castro2018end}, TOPIC~\cite{tao2020few}, SAKD~\cite{cheraghian2021semantic}, SFMS~\cite{cheraghian2021synthesized}, IDLVQ\_C~\cite{chen2021incremental} and Ours on (a) CIFAR100 and (b) \textit{mini}ImageNet.
   }
   %\caption{Example of caption.
   %It is set in Roman so that mathematics (always set in Roman: $B \sin A = A \sin B$) may be included without an ugly clash.}
   \vspace{-0.2in}
   \label{fig:cifar+mini}
\end{figure*}

\section{Experiments}
In this sescion, we first present the dataset and evaluation indicators and then experimental setup. Then, we show the state-of-the-art results of the proposed methods on three datasets with three evaluation indicators. Besides, we conduct comprehensive evaluation on the impacts of Hyper-RPL module, Hyper-Metric module and the related hyper-parameters. Finally, we show the comparison results with more open-set recognition methods.

\subsection{Datasets and Protocols}
We conducted exhaustive experiments on three benchmark datasets for FSCIL: CIFAR100~\cite{krizhevsky2009learning}, \textit{mini}ImageNet~\cite{vinyals2016matching} and CUB200~\cite{chaudhry2018efficient}. We follow the dataset configuration for FSCIL in \cite{tao2020few}, which is illustrated in Table~\ref{table:dataset}.

\begin{itemize}
\item \textbf{CIFAR100}~\cite{krizhevsky2009learning} is widely used in CIL tasks. This dataset involves $100$ categories with $600$ RGB images per class. For each category, $500$ images are applied as the training data and 100 images are the testing data. The size of the image is $32 \times 32$. Specifically, 60 and 40 classes are assigned as the base and novel classes, respectively, and 40 novels classes are learned in a 5-way 5-shot manner for 8 incremental sessions. 

\item \textbf{\textit{mini}ImageNet}~\cite{vinyals2016matching} is constructed by sampling a smaller number of classes from ImageNet. It constitutes $600$ images per class and the image size is $84 \times 84$. We also employ 60 and 40 classes in the base session and incremental learning sessions, respectively. Notably, the incremental sessions is executed with the 5-way 5-shot learning pattern.

\item \textbf{CUB200}~\cite{chaudhry2018efficient} is a fine-grained dataset that contains about $6,000$ training images and $6,000$ test images of over $200$ bird categories. The images are resized
to $256 \times 256$ and then cropped to $224 \times 224$ for training. We apply 100 classes as base classes and split the remaining 100 classes into 10 incremental sessions with the 10-way 5-shot setting.
\end{itemize}

Thorough evaluation on the proposed framework is carried out with three indicators: (1) the final overall accuracy ($\%$) in the last session; (2) the performance dropping rate (PD) ($\%$) \cite{zhang2021few} that measures the absolute accuracy drops in the last session w.r.t. the accuracy in the first session; (3) the average accuracy ($\%$) of all the sessions. 

\subsection{Implementation Details} 

%\noindent
\textbf{Model Configurations.}
For the fair comparison, we employed ResNet-18~\cite{he2016deep} as the backbone for the two branches. After the first session, we froze the parameters of the front four layers of the backbone. During training, the model was optimized by SGD \cite{robbins1951stochastic} (with lr=0.1 and wd=5e-4). For the novel branch, we used the herding selection~\cite{rebuffi2017icarl} when selecting exemplar set for maintain the discriminative competence of previous categories by knowledge distillation. For the classification head, we applied the nearest-mean-of-exemplars (NME) classification. Our framework was implemented using Pytorch  and trained on GeForce RTX 3080 GPUs. 

%\noindent 
\textbf{Training Details.}
For the first session, 160 epochs was executed for the three datasets, which is the same as the traditional FSCIL methods.
For CIFAR100 and \textit{mini}ImageNet, we set the initial learning rate as 0.1, which was divided by 10 after 80 and 120 epochs. For CUB200, the starting learning rate was 0.001, and divided by 10 after 80 and 120 epochs. The model was trained with the training batch size of 128 for \textit{mini}ImageNet and CIFAR100, and 32 for CUB200. As for the integrated distance, $\beta$ and $\gamma$ were assigned as 0.7 and 0.3, respectively. Other parameter configurations followed the setting of RPL. For the incremental sessions, the initial backbone model was pretrained with 100 epochs on the base dataset. For CIFAR100, the model was trained with learning rate 0.01 and batch size 128 for 80 epochs. The Hyper-Metric module was conducted after the fourth session and started from 20 epochs. For \textit{mini}ImageNet, the learning rate was 0.01 and batch size was 128 for 60 epochs. The Hyper-Metric was executed after the fourth session and started from 40 epochs. For CUB200, the learning rate was 0.0005 for the total 60 epochs with training batch size 32. The Hyper-Metric module was performed after the fourth session and started from 20 epochs. The curvature $c$ of the Poincaré ball model is assigned to 0.1. For the hyper-parameters in Equation~\ref{eq:incremental-loss}, $\zeta$ was implemented as the adaptive weight based on the work~\cite{hou2019learning} and $\eta$ was fixed as 1. Lastly, $M$ was equal to the training batch size and $\tau$ is termed 1.

\begin{table*}[t]
\caption{More comparative results with CIFAR100 and \textit{mini}ImageNet. PD is the performance dropping rate from the first session to the last session. Average Acc. is the average performance of all the encountered sessions.}
%Our proposed UaD-CE achieves the state-of-the-art results with respect to the three indicators.
 \vspace{-0.10in}
\centering
\setlength{\tabcolsep}{11pt}
\resizebox{16.0cm}{!}{
\tabcolsep 0.06in
\begin{tabular}{ccccccccccccccccc}
\toprule[1.3pt]
\multirow{2}{*}{\textbf{Dataset}}& \multirow{2}{*}{\textbf{Method}}& \multicolumn{9}{c}{\textbf{Session ID}} & \multirow{2}{*}{\textbf{PD$\downarrow$}} & \multirow{2}{*}{\makecell[c]{\textbf{Average} \\ \textbf{Acc.}}}\\ 
\cline{3-11} 
 & &  \textbf{1} & \textbf{2} &\textbf{3}
 &\textbf{4} & \textbf{5}& \textbf{6}&\textbf{7} & \textbf{8}& \textbf{9} 
\\
\midrule[1.3pt]
\multirow{3}{*}{\textbf{CIFAR100}} & SPPR \cite{zhang2021few} & 76.68 & 72.69 & 67.61 & 63.52 &	59.18 &	55.82 &	53.08 & 50.89 &	48.12 & 28.56 & \textbf{60.68} \\
& CEC~\cite{zhu2021self} & 73.03 &70.86 &65.20 &61.27 &58.03 &55.53& 53.17 &51.19 &49.06 & 23.97 & 59.70 \\
%\cline{2-13}
& \textbf{Ours} & 63.55 & 62.88 & 61.05 & 58.13 & 55.68 & 54.59 & 52.93 &  50.39 & \textbf{49.48} & \textbf{14.07} & 56.52 \\

\midrule[1.3pt]
\multirow{3}{*}{\textbf{\textit{mini}ImageNet}}& SPPR~\cite{zhang2021few} & 80.27 & 74.22 & 68.89 & 64.43 & 60.54 &56.82 &53.81 &	51.22 &	\textbf{48.54} & 31.73& \textbf{62.08} \\
& CEC~\cite{zhu2021self} & 72.22 & 67.06 & 63.17 & 59.79 &56.96 &53.91& 51.36 &49.32 &47.60 & 24.62 & 57.93\\

& \textbf{Ours} & 60.65 & 59.00 & 56.59 & 54.78 & 53.02 & 50.73 & 48.46 & 47.34 & 46.75 & \textbf{13.90} & 53.04\\
\bottomrule[1.3pt]
\end{tabular}
}
\vspace{-0.15in}
\label{table:cifar+mini}
\end{table*}

\begin{table*}[t]
\caption{Comparative study with CUB200. PD is the performance dropping rate from the first session to the last session. Average Acc. is the average performance of all the encountered sessions.}
%Our proposed UaD-CE achieves the state-of-the-art results with respect to the three indicators.
 \vspace{-0.10in}
\centering
\setlength{\tabcolsep}{11pt}
\resizebox{17.0cm}{!}{
\tabcolsep 0.06in
\begin{tabular}{cccccccccccccccc}
\toprule[1.3pt]
\multirow{2}{*}{\textbf{Method}}& \multicolumn{11}{c}{\textbf{Session ID}} & \multirow{2}{*}{\textbf{PD$\downarrow$}} & \multirow{2}{*}{\makecell[c]{\textbf{Average} \\ \textbf{Acc.}}}\\ 
\cline{2-12} 
 &  \textbf{1} & \textbf{2} &\textbf{3}
 &\textbf{4} & \textbf{5}& \textbf{6}&\textbf{7} & \textbf{8}& \textbf{9}& \textbf{10} &\textbf{11} 
\\
\midrule[1.3pt]
Ft-CNN~\cite{tao2020few} & 68.68 & 44.81& 32.26 &25.83 &25.62 &25.22& 20.84 &16.77& 18.82 &18.25 & 17.18 & 51.50 & 28.57\\
Joint-CNN~\cite{tao2020few} & 68.68& 62.43& 57.23& 52.80& 49.50& 46.10& 42.80& 40.10& 38.70& 37.10& 35.60 &  33.08 & 48.28\\
%\hline
iCaRL~\cite{rebuffi2017icarl} &68.68& 52.65& 48.61& 44.16& 36.62& 29.52& 27.83& 26.26& 24.01& 23.89& 21.16 & 47.52 & 36.67\\
EEIL~\cite{castro2018end} &68.68 &53.63 & 47.91 & 44.20& 36.30 & 27.46&25.93& 24.70& 23.95& 24.13& 22.11 &46.57 & 36.27\\
NCM~\cite{hou2019learning}  &68.68 &57.12 & 44.21 &28.78 &26.71 &25.66 &24.62 &21.52 &20.12 &20.06 &19.87 & 48.81 & 32.49\\
%\hline
TOPIC~\cite{tao2020few} &68.68 &62.49 &54.81 &49.99 &45.25 &41.40 &38.35 &35.36&32.22 &28.31 &26.28 & 42.40 & 43.92\\
SAKD~\cite{cheraghian2021semantic} &68.23& 60.45 &55.70& 50.45& 45.72 &42.90& 40.89& 38.77& 36.51& 34.87& 32.96 & 35.27 & 46.13\\
SPPR~\cite{zhu2021self} &68.68& 61.85& 57.43& 52.68& 50.19& 46.88& 44.65 &43.07& 40.17& 39.63& 37.33 & 31.35 & 49.32\\
%IDLVQ-C~\cite{chen2021incremental} & 77.37& 74.72& 70.28& 67.13& 65.34 &63.52 &62.10 &61.54 &59.04& 58.68& 57.81 & 19.56\\
SFMS~\cite{cheraghian2021synthesized} & 68.78& 59.37& 59.32& 54.96& 52.58 &49.81 &48.09 &46.32 &44.33& 43.43& 43.23 & 25.55 & 51.84\\
\midrule[1.3pt]
%ERDIL,2021~\cite{dong2021few} &73.52& 71.09& 66.13& 63.25& 59.49& 59.89& 58.64& 57.72& 56.15& 54.75& 52.28 &21.24 &61.17\\
CEC \cite{zhang2021few} &75.85 &71.94 &68.50 &63.50 &62.43 &58.27& 57.73 &55.81 &54.83 &53.52 &52.28 &23.57 & 61.33\\
IDLVQ-C~\cite{chen2021incremental} & 77.37 &74.72& 70.28& 67.13& 65.34 &63.52& 62.10 &61.54 &59.04& 58.68& \textbf{57.81} & 19.56 & \textbf{65.18} \\
\midrule[1.3pt]
\textbf{Ours} & 63.20 & 62.61 & 59.83 & 56.82 & 55.07 & 53.06 & 51.56 & 50.05 & 47.50 & 46.82 & 45.87 &  \textbf{17.33} & 54.30\\
\bottomrule[1.3pt]
\end{tabular}
}
\vspace{-0.15in}
\label{table:cub200}
\end{table*}

\noindent \subsection{Comparison with State-of-the-art Methods}
We evaluate our proposed framework on three benchmark datasets, and compare it with state-of-the-art methods regarding three indicators. Our comparison is mainly divided into two parts: (1) The first part is to compare with current methods that have the similar classification accuracy with ours in the first session, and we analyse the results towards three evaluation indicators. (2) The other situation is when the classification performance is superior to ours in the first session. Since we endow the model with not only the classification ability of known categories but also the competence of detecting unknown classes, the overall performance can not exceed the single task performance obtained with the traditional FSCIL protocol. Therefore, we demonstrate the dominant performance with respect to PD and final overall accuracy.

\textbf{Results on CIFAR100.} As shown in Figure~\ref{fig:cifar+mini} (a), we compare our method with the current methods that have similar classification accuracy of known categories in the first session. The curve representing our achievement experiences a gentle downtrend, thus the classification accuracy of the last session exceeds all other methods and the PD is the smallest. Meanwhile, this phenomenon also demonstrates that our framework can better tackle the catastrophic forgetting issue. Moreover, the curve representing our framework is above other curves, which means that the surpassing average accuracy is achieved by the proposed framework. Table~\ref{table:cifar+mini} (Rows 3-5) presents more comparative results. As for the indicator of average accuracy, the results obtained by CEC~\cite{zhang2021few} and SPPR~\cite{zhu2021self} are better than ours, which results from the predominant performance of the first session. It can be concluded that our framework achieves the superior performance toward three evaluation indicators on CIFAR100.

\textbf{Results on \textit{mini}ImageNet.} Figure~\ref{fig:cifar+mini} (b) compares the performance
of our method with state-of-the-art methods. For the first session, the open-set recognition task is endowed to the model for leaving the possibility of novel encountered classes, which exists the trade-off issue of the performance between known and unknown classes. Therefore, the classification accuracy on known classes can not exceed other methods that only conduct the classification task on known classes. Therefore, it can be explained that why our framework obtains the inferior performance to other methods in the first two sessions. For the third session, our proposed framework outpaces other methods, and then goes through a slight decline. Table~\ref{table:cifar+mini} (Rows 6-8) compares the performance of our method with CEC~\cite{zhang2021few} and SPPR~\cite{zhu2021self}, and Our framework achieves less PD than these two methods.

\textbf{Results on CUB200.} As illustrated in Table~\ref{table:cub200} (Rows 3-11), when the model possesses the similar classification performance in the first session, our framework reaches the superior performance in regard to the three evaluation indicators.
Table~\ref{table:cub200} (Rows 12-13) shows the comparative results with CEC~\cite{zhang2021few} and IDLVQ-C~\cite{chen2021incremental}. Specifically, our proposed framework suffers from less performance dropping rate (17.33\%), which reflects the outstanding ability of memorizing the old knowledge.

\begin{table*}[t]
\caption{The efficacy of Hyper-RPL module.}
%Our proposed UaD-CE achieves the state-of-the-art results with respect to the three indicators.
 \vspace{-0.10in}
\centering
\setlength{\tabcolsep}{11pt}
\resizebox{18.0cm}{!}{
\tabcolsep 0.06in
\begin{tabular}{ccccccccccccccccc}
\toprule[1.3pt]
\multirow{2}{*}{\textbf{Dataset}}& \multirow{2}{*}{\textbf{Threshold}}& \multirow{2}{*}{\textbf{Method}}& \multicolumn{11}{c}{\textbf{Session ID}} & \multirow{2}{*}{\textbf{PD$\downarrow$}} & \multirow{2}{*}{\makecell[c]{\textbf{Average} \\ \textbf{Acc.}}}\\ 
\cline{4-14} 
 & & & \textbf{1 (Known/Unknown)} & \textbf{2} &\textbf{3}
 &\textbf{4} & \textbf{5}& \textbf{6}&\textbf{7} & \textbf{8}& \textbf{9} & \textbf{10} & \textbf{11}
\\
\midrule[1.3pt]
\multirow{2}{*}{\textbf{CIFAR100}} & \multirow{2}{*}{0.75} & RPL & 63.33 / 67.85 & 62.51 & 60.60 & 57.67 & 55.20 & 54.05 & 52.37 & 49.85 & 48.91 & -& - & 14.42 & 56.05 \\
%\cline{2-13}
& & Hyper-RPL \textbf{(Ours)} & \textbf{63.55 / 70.25} & 62.88 & 61.05 & 58.13 & 55.68 & 54.59 & 52.93 & 50.39 & \textbf{49.48} &- & - & \textbf{14.07} & \textbf{56.52} \\

\midrule[1.3pt]
\multirow{2}{*}{\textbf{\textit{mini}ImageNet}}& \multirow{2}{*}{0.73} & RPL & 59.83 / 68.35 & 58.12 & 55.70 & 53.86 & 52.08 & 49.82 & 47.57 & 46.43 & 45.54 & - & - & 14.29 &  52.10 \\

& & Hyper-RPL \textbf{(Ours)} & \textbf{60.65 / 71.38} & 59.00 & 56.59 & 54.78 & 53.02 & 50.73 & 48.46 & 47.34 & \textbf{46.75} & - &- & \textbf{13.90} & \textbf{53.04} \\
\midrule[1.3pt]
\multirow{2}{*}{\textbf{CUB200}}& \multirow{2}{*}{0.73} & RPL & 63.09 / 69.83 & 62.26 & 59.38 & 56.29 & 54.50 & 52.45 & 50.90 & 49.36 & 46.84 & 46.26 &  45.10 & 17.99 & 53.31 \\

& & Hyper-RPL \textbf{(Ours)} & \textbf{63.20 / 73.38} & 62.61 & 59.83 & 56.82 & 55.07 & 53.06 & 51.56 & 50.05 & 47.50 & 46.82 & \textbf{45.87} & \textbf{17.33} & \textbf{54.30}\\ 

\bottomrule[1.3pt]
\end{tabular}
}
\vspace{-0.025in}
\label{table:ablation_hyper_RPL}
\end{table*}

\begin{table*}[t]
\caption{The efficacy of Hyper-Metric module. \textit{E} and \textit{H} mean that the operation is conducted in the Euclidean space and hyperbolic space, respectively}
%Our proposed UaD-CE achieves the state-of-the-art results with respect to the three indicators.
\vspace{-0.10in}
\centering
\setlength{\tabcolsep}{11pt}
\resizebox{18.0cm}{!}{
\tabcolsep 0.06in
\begin{tabular}{ccccccccccccccccc}
\toprule[1.3pt]
\multirow{2}{*}{\textbf{Dataset}}& \multirow{2}{*}{$\mathcal{L}^{ce}$}& \multirow{2}{*}{$\mathcal{L}^{dl}$} & \multirow{2}{*}{$\mathcal{L}^m$} & \multicolumn{11}{c}{\textbf{Session ID}} & \multirow{2}{*}{\textbf{PD$\downarrow$}} & \multirow{2}{*}{\makecell[c]{\textbf{Average} \\ \textbf{Acc.}}}\\ 
\cline{5-15} 
& & & &  \textbf{1} & \textbf{2} &\textbf{3}
 &\textbf{4} & \textbf{5}& \textbf{6}&\textbf{7} & \textbf{8}& \textbf{9} & \textbf{10} & \textbf{11}
\\
\midrule[1.3pt]
\multirow{12}{*}{\textbf{CIFAR100}} & \textit{E} & \textit{E} & \XSolidBrush & 63.55 & 62.84 & 60.70 & 56.65 & 55.34 & 54.03 & 52.29 & 49.94 & 49.00 & -&-& 14.55& 56.03 \\
& \textit{H} & \textit{E} &  \XSolidBrush & 63.55 & 62.69 & 60.33 & 56.75 & 53.90 & 52.07 & 50.34 & 47.84 & 46.93 & -& -& 16.62 & 54.93\\
& \textit{E} & \textit{H} &  \XSolidBrush & 63.55 & 62.76 & 60.46 & 57.06 & 54.30 & 52.42& 50.56 & 48.18 & 47.37 & - &-&16.18& 55.18\\
& \textit{H} & \textit{H} &  \XSolidBrush & 63.55 &62.71 & 60.32 & 56.72 & 53.82 & 52.00 & 50.33 & 47.85 & 46.90 & - & - & 16.65 & 54.91\\
\cline{2-17}
& \XSolidBrush & \textit{H} & \textit{E} & 63.55 & 62.65 & 60.10 & 56.20 & 53.63 & 51.49 & 49.58 & 46.62 & 45.61 & - & - & 17.94 & 54.38 \\
& \XSolidBrush & \textit{E} & \textit{E} & 63.55 & 62.60 & 60.11 & 56.42 & 53.43 & 51.72 & 50.03 & 47.17 & 46.50 & - & - & 17.05 & 54.61\\
& \textit{H} & \textit{H} & \textit{E} & 63.55 & 62.62 & 60.13 & 56.17 & 53.61 & 51.54 & 50.05 & 47.22 & 45.01 & - & - & 18.53 & 54.43 \\
& \textit{H} & \textit{E} & \textit{E} & 63.55 & 62.56 & 60.10 & 56.62 & 53.36 & 51.59 & 49.77 & 47.14 & 46.29 & - & - & 17.26 & 54.55 \\
& \textit{E} & \textit{H} & \textit{E} & 63.55 & 62.56 & 60.23 & 56.24 & 53.53 & 51.74 & 49.98 & 47.18 & 46.43 & - & - & 17.12 & 54.60 \\
& \textit{E} & \textit{E} & \textit{E} & 63.55 & 62.71 & 60.19 & 57.15 & 53.47 & 52.78 & 51.45 & 48.81 & 48.06 & - & - & 15.49 & 55.35 \\
\cline{2-17}
%\cline{2-13}
& \XSolidBrush & \textit{H} & \textit{H} & 63.55 & 62.67 & 60.40 & 56.67 & 53.98 & 52.14 & 50.39 & 47.87 & 46.80 & - & - & 16.75 & 54.94 \\
& \XSolidBrush & \textit{E} & \textit{H} & 63.55 & 62.69 & 60.31 & 56.71 & 53.97 & 52.06 & 50.38 & 47.67 & 46.75 & - & - & 16.80 & 54.90 \\
& \textit{H} & \textit{H} & \textit{H} & 63.55 & 62.68 & 60.48 & 56.70 & 54.00 & 51.98 & 50.35 & 46.63 & 46.78 & - & - &  16.77 & 54.79\\
& \textit{H} & \textit{E} & \textit{H} & 63.55 & 62.71 & 60.36 & 56.67 & 53.90 & 51.89 & 50.26 & 47.56 & 46.66 & - & - & 16.89 & 54.84\\
& \textit{E} & \textit{H} & \textit{H} & 63.55 & 62.73 & 60.39 & 56.75 & 54.13 & 52.10 & 50.55 & 47.82 & 46.81 & - & - & 16.74 & 54.98\\
& \textit{E} & \textit{E} & \textit{H} & 63.55 & 62.88 & 61.05 & 58.13 & 55.68 & 54.59 & 52.93 & 50.39 & \textbf{49.48} &- & - & \textbf{14.07} & \textbf{56.52} \\

\midrule[1.3pt]
\multirow{3}{*}{\textbf{\textit{mini}ImageNet}}& \textit{E} & \textit{E} & \XSolidBrush & 60.65 & 58.99 & 56.48 & 54.60 & 52.90 & 50.63 & 48.35 & 46.96 & 46.49 & - & - & 14.16 & 52.89 \\

& \textit{E} & \textit{E} & \textit{E} & 60.65 & 59.04 & 56.51 & 53.92 & 52.21 & 49.71 & 47.63 & 46.35 & 42.76 & - & -  & 17.89 & 52.09 \\

& \textit{E} & \textit{E} & \textit{H} & 60.65 & 59.00 & 56.59 & 54.78 & 53.02 & 50.73 & 48.46 & 47.34 & \textbf{46.75} & - &- & \textbf{13.90} & \textbf{53.04} \\
\midrule[1.3pt]
\multirow{3}{*}{\textbf{CUB200}}& \textit{E} & \textit{E} & \XSolidBrush & 63.20 & 62.44 & 59.23 & 56.29 &54.46 & 52.46 & 50.92 & 49.81 & 47.80 & 46.43 & 45.72 & 17.48 & 53.52  \\

& \textit{E} & \textit{E} & \textit{E} & 63.20 & 62.42 & 59.41 & 55.93 & 54.66 & 51.90 & 50.83 & 50.02 & 47.50 & 47.62& 45.71 & 17.49 & 53.56 \\

& \textit{E} & \textit{E} & \textit{H} & 63.20 & 62.61 & 59.83 & 56.82 & 55.07 & 53.06 & 51.56 & 50.05 & 47.50 & 46.82 & \textbf{45.87} & \textbf{17.33} & \textbf{54.30}\\ 

\bottomrule[1.3pt]
\end{tabular}
}
\vspace{-0.1in}
\label{table:ablation_hyper_metric}
\end{table*}

\begin{figure}[t]
  \centering
  %\fbox{\rule{0pt}{2in} \rule{0.9\linewidth}{0pt}}
   \includegraphics[width=0.8\linewidth]{ 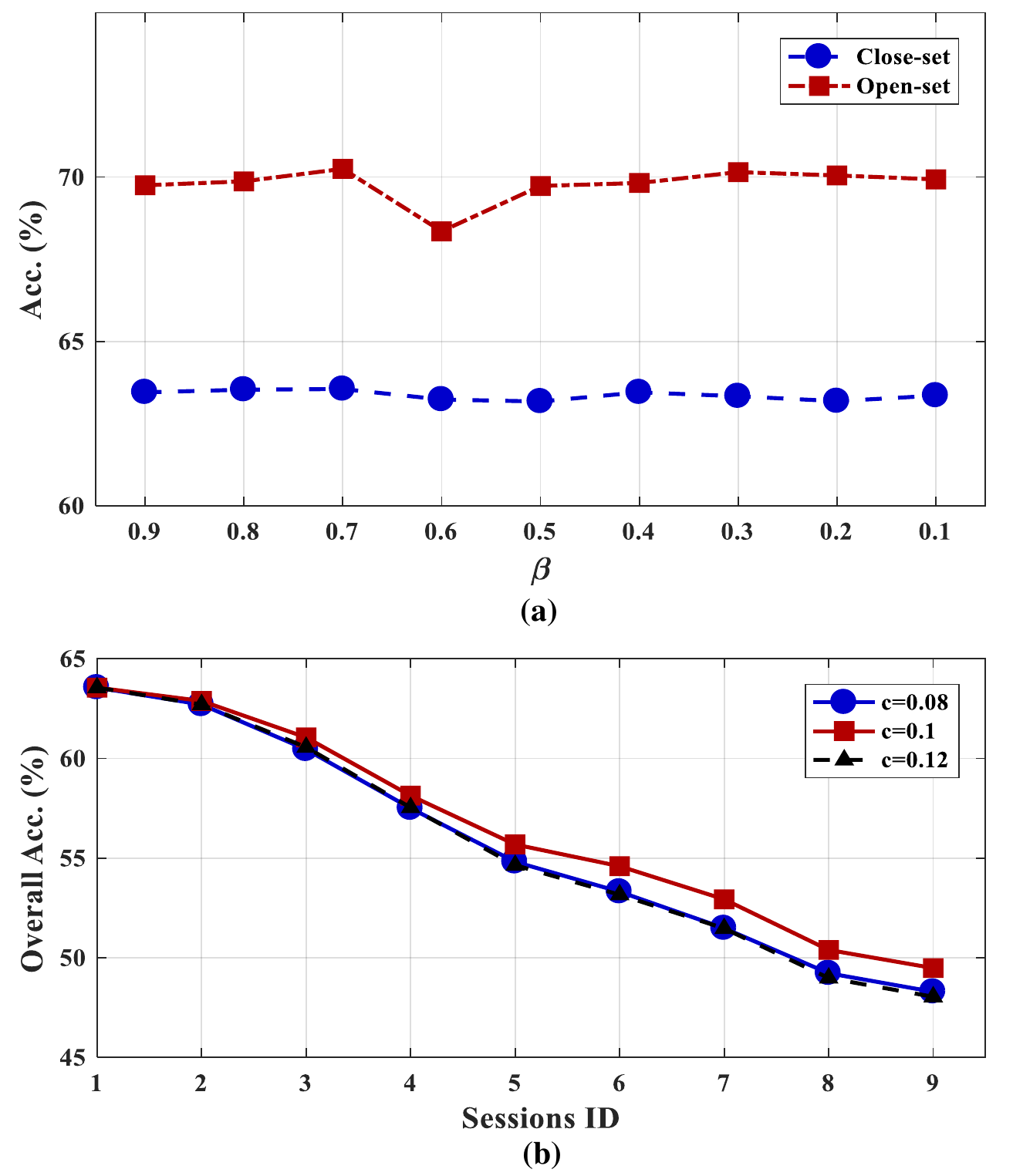}
   \vspace{-0.15in}
   \caption{Ablation results on the (a) distance weights; and (b) curvature for the Poincaré ball model. For (a), $\gamma = 1 - \beta$, so we only list the value of $\gamma$ here.}
   
   %$\beta$ and $\gamma$ ($\gamma$=1-$\beta$)
   
   %\caption{Example of caption.
   %It is set in Roman so that mathematics (always set in Roman: $B \sin A = A \sin B$) may be included without an ugly clash.}
   \vspace{-0.2in}
   \label{fig:ablation}
\end{figure}

\begin{table}[t]
\caption{Ablation study on temperature $\tau$ in Hyper-Metric module.}
%Our proposed UaD-CE achieves the state-of-the-art results with respect to the three indicators.
 \vspace{-0.10in}
\centering
\setlength{\tabcolsep}{11pt}
\resizebox{9.0cm}{!}{
\tabcolsep 0.06in
\begin{tabular}{cccccccccccccccc}
\toprule[1.3pt]
\multirow{2}{*}{\large$\tau$}& \multicolumn{9}{c}{\textbf{Session ID}} & \multirow{2}{*}{\textbf{PD$\downarrow$}} & \multirow{2}{*}{\makecell[c]{\textbf{Average} \\ \textbf{Acc.}}}\\ 
\cline{2-10} 
 &  \textbf{1} & \textbf{2} &\textbf{3}
 &\textbf{4} & \textbf{5}& \textbf{6}&\textbf{7} & \textbf{8}& \textbf{9} 
\\
\midrule[1.3pt]
0.7 & 63.55 & 62.75 & 60.39 & 57.43 & 55.03 & 53.15& 51.14 & 48.44 & 47.87 & 15.68 & 55.53 \\
0.8 & 63.55 & 62.70 & 60.44 & 57.33 & 54.88 & 53.10 & 51.55 & 49.15 & 48.53 & 15.02 & 55.69  \\
0.9 & 63.55 & 62.74 & 60.57 & 57.52 & 55.01 & 53.53 & 51.87 & 49.28& 48.97 & 14.58 & 55.89 \\

\textbf{1.0} & 63.55 & 62.88 & 61.05 & 58.13 & 55.68 & 54.59 & 52.93 & 50.39 & \textbf{49.48} & \textbf{14.07} & \textbf{56.52} \\
\bottomrule[1.3pt]
\end{tabular}
}
\vspace{-0.20in}
\label{table:ablation_temperature}
\end{table}

\begin{table*}[t]
\caption{Prediction accuracy novel classes on three benchmark datasets.}
\vspace{-0.1in}
\centering
\setlength{\tabcolsep}{11pt}
\resizebox{16.0cm}{!}{
\tabcolsep 0.06in
\begin{tabular}{cccccccccccccccc}
\toprule[1.3pt]
\multirow{2}{*}{\textbf{Dataset}}& \multirow{2}{*}{\textbf{Method}} & \multirow{2}{*}{\makecell[c]{\textbf{Classes}}}& \multicolumn{9}{c}{\textbf{Session ID}}\\ 
\cline{4-14} 
&  &   & \textbf{1} & \textbf{2} &\textbf{3}
 &\textbf{4} & \textbf{5}& \textbf{6}&\textbf{7} & \textbf{8}& \textbf{9} &\textbf{10}& \textbf{11}
\\
\midrule[1.3pt]
\multirow{6}{*}{\rotatebox{90}{\textbf{CIFAR100}}}& 
%\multirow{3}{*}{SPPR,2021~\cite{zhu2021self}}& Base  & 76.68 & 77.48
%& 77.32 & 76.95	&60.25 & 76.42 &75.95 & 75.83 &  75.30 &-&-\\
\multirow{2}{*}{SPPR~\cite{zhu2021self}} &Novel & - & 1.52 & 9.40	 & 9.80	& 7.45 & 7.52 & 7.33 &  8.14 & 7.35&-&- \\
& &  All & 76.68 & 72.69 &	67.61 &	63.52 &	59.18 &	55.82 &	53.08 & 50.89 &	48.12 &-&-\\
\cline{2-14}
%& \multirow{3}{*}{CEC,2021 \cite{zhang2021few}}  
%&Base & 73.03 & 73.77
% & 71.52	 &70.78	& 70.07&69.28	 &68.70 	 & 68.45 &67.78&-&-\\
& \multirow{2}{*}{CEC \cite{zhang2021few}} &Novel & - & 36.00
& 27.30	 & 22.67 & 21.90 & 22.52 & 22.10	 & 21.60 &  20.98&-&- \\
& &All&73.03 &70.86 &65.20 &61.27 &58.03 &55.53& 53.17 &51.19 &49.06&-&- \\
\cline{2-14}
& \multirow{2}{*}{Ours} %& Base & 63.55 &63.55&63.55&63.55&63.55&63.55&63.55&63.55 & 63.55 & - & -\\
&Novel & - & \textbf{54.79} & \textbf{46.03} & \textbf{36.43} & \textbf{32.05} & \textbf{33.07} & \textbf{31.70} & \textbf{27.84} & \textbf{28.38} &-&- \\
& & All & 63.55 & 62.88 & 61.05 & 58.13 & 55.68 & 54.59 & 52.93 &  50.39 & 49.48 & - & -\\

\bottomrule[1.3pt]
\multirow{6}{*}{\textbf{\rotatebox{90}{\textit{mini}ImageNet}}}& \multirow{2}{*}{SPPR~\cite{zhu2021self}}  
%&Base & 80.27& 80.22 & 80.28 & 80.20 & 80.08 &	79.77 &	79.98 &	79.58 &	79.87&-&-\\
&Novel & - &2.20  & 0.50
& 1.33	 &1.90	& 1.76&1.47	 &	2.60 & 1.55&-&-   \\
& & All & 80.27 & 74.22 & 68.89 & 64.43 & 60.54 &56.82 &53.81 & 51.22 & 48.54&-&-\\
\cline{2-14}
& \multirow{2}{*}{CEC \cite{zhang2021few}} %& Base  & 72.22 &70.92 & 70.17 &69.65	& 69.32&68.98 	 &	68.68 & 68.25 &67.87 & - & - \\
&Novel &-  & 20.80 &21.20 
& 20.33	 &19.90	& 17.72&16.70&16.86&17.20&-&-\\
& & All & 72.22 & 67.06 & 63.17 & 59.79 &56.96 &53.91& 51.36 &49.32 &47.60 &-&-\\

\cline{2-14}
% base 60.65  openset 71.38
 & \multirow{2}{*}{Ours} &Novel & - & \textbf{39.26} & \textbf{32.26} & \textbf{31.31} & \textbf{30.12} & \textbf{26.95} & \textbf{24.08} &  \textbf{24.51} & \textbf{25.89} & - & - \\
& &All & 60.65 & 59.00 & 56.59 & 54.78 & 53.02 & 50.73 & 48.46 & 47.34 & 46.75 &-& - \\

\bottomrule[1.3pt]
\multirow{6}{*}{\rotatebox{90}{\textbf{CUB200}}} &  \multirow{2}{*}{SPPR~\cite{zhu2021self}}  
%&Base & 68.16 & 60.67 & 59.87 & 60.13 & 57.82 & 55.85& 53.77&	53.57 & 52.95& 52.43 & 51.73 \\
& Novel  & -  & \textbf{60.39} & \textbf{45.55} & 30.24 & 31.07 & 27.75 & 26.67& 27.51	 & 25.16 & 24.76&  23.88 \\
 & & All & 68.16& 60.32& 57.11&52.79&49.68&45.95& 43.19 & 42.39 & 40.17 & 38.93 & 37.33 \\
\cline{3-14}
&  \multirow{2}{*}{CEC \cite{zhang2021few}} 
%& Base   & 75.82& 74.34& 73.94 & 73.59 & 72.73 & 72.39 & 71.90& 71.28	 & 71.14	 & 70.90 &  70.66\\
&Novel  & - & 46.61 & 41.74 & 33.34 & \textbf{37.46} & 32.77 &	\textbf{34.78} & \textbf{35.06} & \textbf{33.23} & \textbf{34.98} & \textbf{34.17} \\
& &All & 75.82 & 71.91& 68.52&63.53& 62.45& 58.27 & 57.62& 55.81& 54.85&53.52&52.26 \\
\cline{3-14}
% base 63.20 openset 73.38
&  \multirow{2}{*}{Ours}  &Novel  & - & 56.81 & 43.17 & \textbf{35.16} & 34.99 & \textbf{33.03} & 32.41& 31.51& 28.13& 28.86 & 28.73 \\
& &All & 63.20 & 62.61 & 59.83 & 56.82 & 55.07 & 53.06 & 51.56 & 50.05 & 47.50 & 46.82 & 45.87\\

\bottomrule[1.3pt]
\end{tabular}
}
\vspace{-0.1in}
\label{table:acc}
\end{table*}

\subsection{Ablation Study}
\textbf{Efficacy of Hyper-RPL module.}
Here we evaluate the efficacy of Hyper-RPL module with the combination of hyperbolic geometry to provide a more comprehensive searching space, we present the comparative result with RPL in Table~\ref{table:ablation_hyper_RPL}. When we assign the same threshold to RPL and Hyper-RPL for detecting the unknown categories, the classification accuracy of known categories for these two methods is similar, while Hyper-RPL performs better on detecting the unknown categories. Therefore, with Hyper-RPL, more novel samples can be classified as unknown categories and sent to the novel branch, which results in the surpassing performance of incremental sessions. Furthermore, it is worth noting that the incorporated searching space can assist the model to find a better decision boundary for known and unknown categories.

\begin{figure*}[t]
  \centering
  %\fbox{\rule{0pt}{2in} \rule{0.9\linewidth}{0pt}}
   \includegraphics[width=1.0\linewidth]{ 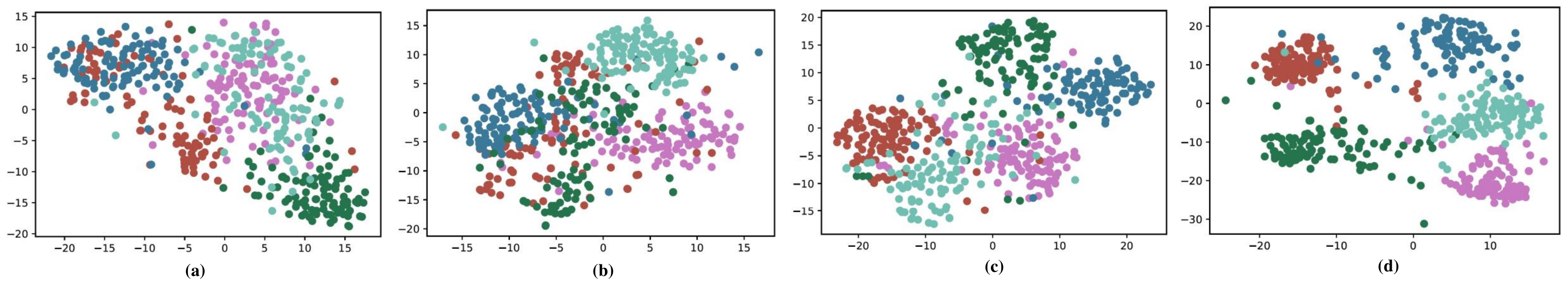}
   \vspace{-0.3in}
   \caption{t-SNE results of CIFAR100 in the 7-th session. Different colors indicate features from five novel classes: \textit{green}=seal, \textit{blue}=television, \textit{mint green}=tiger, \textit{pink}=table and \textit{orange}=street\_car. The t-SNE results correspond to the four cases in Table~\ref{table:ablation_hyper_metric}: (a) \textit{H}, \textit{E}, \textit{E}; (b) \textit{E}, \textit{E}, \textit{E}; (c) \textit{E}, \textit{E}, \XSolidBrush; (d) \textit{E}, \textit{E}, \textit{H}.
   }
   %\caption{Example of caption.
   %It is set in Roman so that mathematics (always set in Roman: $B \sin A = A \sin B$) may be included without an ugly clash.}
   \vspace{-0.1in}
   \label{fig:tsne}
\end{figure*}

\textbf{Impact of integrated distance weights.} In these experiments, we evaluate the impacts of distance weights $\beta$ and $\gamma$ in Equation~\ref{eq:d_all} by assigning different values. 
In Hyper-Metric module, we introduce the hyperbolic geometry to the current Euclidean space, and the novel integrated function (\ie, Equation~\ref{eq:d_all}) is proposed. The integrated distance is used for computing the logits which then generate the predictions of the model. Therefore, the distance weights $\beta$ and $\gamma$ determines the searching space for optimizing the model. The ablation study results are illustrated in Figure~\ref{fig:ablation} (a), and Hyper-RPL module achieves the superior discriminative ability of detecting open-set categories when $\beta$ and $\gamma$ are assigned 0.7 and 0.3, respectively.

\textbf{Comparison with more open-set recognition methods.} To further validate the efficacy of Hyper-metric module, we compare it with two other open-set recognition methods:  Openmax~\cite{bendale2016towards} and PROSER~\cite{zhou2021learning}. Openmax belongs to the discriminative model-based method, and the close-set and open set recognition accuracy are 45.30\% and 59.52\%, respectively. PROSER is based on the generative model, the classification accuracy of seen and unseen categories are 63.30\% and 69.52\%, respectively. RPL can be also regarded as discriminative model-based method, while the novelty of formulating close-set recognition loss and the open-space risk contribute to the final excellent performance. By integrating the hyperbolic geometry in Hyper-RPL, the clearer decision boundary between seen and unseen categories is founded with the extended optimization space.

\textbf{Efficacy of Hyper-Metric module.} We present this ablation study results in Table~\ref{table:ablation_hyper_metric}. $\mathcal{L}^{ce}$, $\mathcal{L}^{dl}$ and $\mathcal{L}^{dl}$ are the three components of Equation~\ref{eq:incremental-loss}. `\textit{E}' and `\textit{H}' represent that the corresponding loss is operated in Euclidean space and hyperbolic space, respectively. `\XSolidBrush' means that the loss is not included into the optimization process. We use the third row as an example, and the case is that we do not combine the metric learning loss in Equation~\ref{eq:incremental-loss}, \ie, optimizing the model with cross entropy classification loss and distillation loss in Euclidean space. We conduct ablation experiments on all possible cases. Since novel categories are all in the limited data regime, the learning algorithm easily suffers from the overfitting issue. If the metric learning is carried out in Euclidean space which is the same as the other two losses, the performance is deteriorated, which is due to the much severe overfitting issue. However, when the metric learning is executed in hyperbolic space and the other losses are executed in Euclidean space, the performance enhancement is obtained since we provide a more abundant learning space to alleviate the overfitting issue. 

\textbf{Impact of the temperature $\tau$ in Hyper-Metric module.} The value of the temperature in Equation~\ref{eq:metric_loss} is usually less than 1, and we increase this number from 0.7 to 1 to evaluate the impact of this hyper-parameter. The ablation study results are shown in Table~\ref{table:ablation_temperature}. It can be concluded that the performance improves as temperature increases owing to the more smooth logits, and achieves the best when the value is assigned to 1.

\textbf{Impact of the curvature $c$ for the Poincaré ball model.} We mainly evaluate the impact of the curvature $c$ on the incremental learning session, so we fix the performance of the first session. As shown in Figure~\ref{fig:ablation} (b), our proposed framework achieves the best overall performance when the curvature of the Poincaré ball model is assigned to 0.1. the curvature $c$ is usually regarded as the balance parameter between hyperbolic and Euclidean geometries, which means that this hyper-parameter also affect the integrated optimization space that the model can access. The performance of our framework is undermined when the value of $c$ increases or decreases.

\begin{figure}[t]
  \centering
  %\fbox{\rule{0pt}{2in} \rule{0.9\linewidth}{0pt}}
   \includegraphics[width=0.75\linewidth]{ 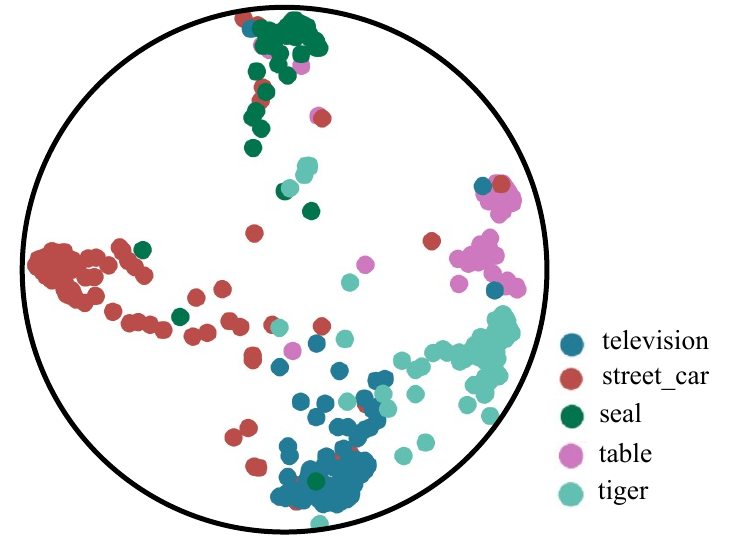}
   \vspace{-0.1in}
   \caption{Visualization in Poincaré ball model corresponding to the situation shown in Figure~\ref{fig:tsne} (d). 
   }
   %\caption{Example of caption.
   %It is set in Roman so that mathematics (always set in Roman: $B \sin A = A \sin B$) may be included without an ugly clash.}
   \vspace{-0.2in}
   \label{fig:hyper-visualization}
\end{figure}

\subsection{Further Analysis and Visualization}
In our proposed framework, the classification performance of the first session can be totally preserved. Furthermore, we show the classification accuracy of novel categories in Table~\ref{table:acc}. It can be concluded that our proposed framework can better alleviate the overfitting issue than CEC~\cite{zhang2021few} and SPPR~\cite{zhu2021self} to a large extent. This achievement result from two aspects: (1) The trade-off issue is better mitigated by the proposed novel framework for FSCIL. (2) The Hyper-Metric module extends the optimization space for learning novel categories to alleviate the overfitting issue.

Figure \ref{fig:tsne} illustrates the t-SNE results of CIFAR100 in the 7-th session. For Figure \ref{fig:tsne} (a) and Figure \ref{fig:tsne} (b), features of the same category are more concentrated with the cross-entropy loss in Euclidean space, which means that the intra-class difference is reduced. From For Figure \ref{fig:tsne} (b) and Figure \ref{fig:tsne} (c), the decision boundaries for the five classes become clearer when we remove the metric-learning operation in Euclidean space. When metric learning is executed in Euclidean geometry, the overfitting issue becomes severer due to the extra optimization process with limited labeled data. However, with the Hyper-Metric module, our proposed framework achieves the surpassing discriminative ability. Hyper-Metric module integrates the hyperbolic geometry into the existing Euclidean geometry, which provides a more comprehensive searching space for the optimization process. Furthermore, the visualization in Poincaré ball model corresponding to the situation shown in Figure~\ref{fig:tsne} (d) is shown in Figure~\ref{fig:hyper-visualization}. Like in Euclidean space, the representation in hyperbolic geometry is also discriminative. Besides, most of samples locate near the boundary of the circle, which means that the prediction uncertainty is low.

\section{Conclusion and Future Work}
Currently, the protocol of FSCIL is built by imitating the general CIL setting which is not totally appropriate due to the different data configurations. In this paper, we rethink the FSCIL with open-set hypothesis and propose a more practical configuration.
To realize the better discriminative ability of both seen categories and unseen categories in the first session, Hyper-RPL is put forward by incorporating hyperbolic geometry to obtain a clearer decision boundary. Moreover, Hyper-Metric is introduced into the distillation-based CIL framework to handle the overfitting and trade-off issues, which results from a more comprehensive learning space. The exhaustive evaluations of the proposed configuration and modules on three benchmark datasets are conducted to certify the efficacy regarding three evaluation indicators.

In spite of the remarkable performance achieved by the proposed framework, two future improvement directions still include: (1) Since our framework benefits from introducing hyperbolic geometry to existing Euclidean geometry, the integrated strategy is critical for the final achievements. We will seek for the adaptive integrated strategy to replace the current well-designed one in the future. (2) The hyperbolic geometry is already a rich representation space, however, it can be seen from Table~\ref{table:ablation_hyper_metric} that our method can not achieve satisfying results when only hyperbolic geometry is applied. Thus, we will explore valid solutions to operate FSCIL task in this single space.

\ifCLASSOPTIONcaptionsoff
  \newpage
\fi

\normalem
\bibliographystyle{IEEEtran}
\bibliography{references}

% Generated by IEEEtran.bst, version: 1.12 (2007/01/11)
\begin{thebibliography}{10}
\providecommand{\url}[1]{#1}
\csname url@samestyle\endcsname
\providecommand{\newblock}{\relax}
\providecommand{\bibinfo}[2]{#2}
\providecommand{\BIBentrySTDinterwordspacing}{\spaceskip=0pt\relax}
\providecommand{\BIBentryALTinterwordstretchfactor}{4}
\providecommand{\BIBentryALTinterwordspacing}{\spaceskip=\fontdimen2\font plus
\BIBentryALTinterwordstretchfactor\fontdimen3\font minus
  \fontdimen4\font\relax}
\providecommand{\BIBforeignlanguage}[2]{{%
\expandafter\ifx\csname l@#1\endcsname\relax
\typeout{** WARNING: IEEEtran.bst: No hyphenation pattern has been}%
\typeout{** loaded for the language `#1'. Using the pattern for}%
\typeout{** the default language instead.}%
\else
\language=\csname l@#1\endcsname
\fi
#2}}
\providecommand{\BIBdecl}{\relax}
\BIBdecl

\bibitem{zenke2017continual}
F.~Zenke, B.~Poole, and S.~Ganguli, ``Continual learning through synaptic
  intelligence,'' in \emph{ICML}, 2017, pp. 3987--3995.

\bibitem{lopez2017gradient}
D.~Lopez-Paz and M.~Ranzato, ``Gradient episodic memory for continual
  learning,'' \emph{Advances in neural information processing systems},
  vol.~30, 2017.

\bibitem{snell2017prototypical}
J.~Snell, K.~Swersky, and R.~Zemel, ``Prototypical networks for few-shot
  learning,'' in \emph{NeurIPS}, 2017, pp. 4080--4090.

\bibitem{finn2017model}
C.~Finn, P.~Abbeel, and S.~Levine, ``Model-agnostic meta-learning for fast
  adaptation of deep networks,'' in \emph{ICML}, 2017, pp. 1126--1135.

\bibitem{cui2022coarse}
Y.~Cui, Q.~Liao, D.~Hu, W.~An, and L.~Liu, ``Coarse-to-fine pseudo supervision
  guided meta-task optimization for few-shot object classification,''
  \emph{Pattern Recognition}, vol. 122, p. 108296, 2022.

\bibitem{tao2020few}
X.~Tao, X.~Hong, X.~Chang, S.~Dong, X.~Wei, and Y.~Gong, ``Few-shot
  class-incremental learning,'' in \emph{CVPR}, 2020, pp. 12\,183--12\,192.

\bibitem{van2019three}
G.~M. Van~de Ven and A.~S. Tolias, ``Three scenarios for continual learning,''
  \emph{arXiv preprint arXiv:1904.07734}, 2019.

\bibitem{hou2018lifelong}
S.~Hou, X.~Pan, C.~C. Loy, Z.~Wang, and D.~Lin, ``Lifelong learning via
  progressive distillation and retrospection,'' in \emph{ECCV}, 2018, pp.
  437--452.

\bibitem{rebuffi2017icarl}
S.-A. Rebuffi, A.~Kolesnikov, G.~Sperl, and C.~H. Lampert, ``icarl: Incremental
  classifier and representation learning,'' in \emph{CVPR}, 2017, pp.
  2001--2010.

\bibitem{zhang2021few}
C.~Zhang, N.~Song, G.~Lin, Y.~Zheng, P.~Pan, and Y.~Xu, ``Few-shot incremental
  learning with continually evolved classifiers,'' in \emph{CVPR}, 2021, pp.
  12\,455--12\,464.

\bibitem{zhu2021self}
K.~Zhu, Y.~Cao, W.~Zhai, J.~Cheng, and Z.-J. Zha, ``Self-promoted prototype
  refinement for few-shot class-incremental learning,'' in \emph{CVPR}, 2021,
  pp. 6801--6810.

\bibitem{scheirer2012toward}
W.~J. Scheirer, A.~de~Rezende~Rocha, A.~Sapkota, and T.~E. Boult, ``Toward open
  set recognition,'' \emph{IEEE transactions on pattern analysis and machine
  intelligence}, vol.~35, no.~7, pp. 1757--1772, 2012.

\bibitem{geng2020recent}
C.~Geng, S.-j. Huang, and S.~Chen, ``Recent advances in open set recognition: A
  survey,'' \emph{IEEE transactions on pattern analysis and machine
  intelligence}, vol.~43, no.~10, pp. 3614--3631, 2020.

\bibitem{chen2020learning}
G.~Chen, L.~Qiao, Y.~Shi, P.~Peng, J.~Li, T.~Huang, S.~Pu, and Y.~Tian,
  ``Learning open set network with discriminative reciprocal points,'' in
  \emph{European Conference on Computer Vision}.\hskip 1em plus 0.5em minus
  0.4em\relax Springer, 2020, pp. 507--522.

\bibitem{peng2021hyperbolic}
W.~Peng, T.~Varanka, A.~Mostafa, H.~Shi, and G.~Zhao, ``Hyperbolic deep neural
  networks: A survey,'' \emph{IEEE Transactions on Pattern Analysis \& Machine
  Intelligence}, no.~01, pp. 1--1, 2021.

\bibitem{schwartz2018delta}
E.~Schwartz, L.~Karlinsky, J.~Shtok, S.~Harary, M.~Marder, A.~Kumar, R.~Feris,
  R.~Giryes, and A.~Bronstein, ``Delta-encoder: an effective sample synthesis
  method for few-shot object recognition,'' \emph{Advances in neural
  information processing systems}, vol.~31, 2018.

\bibitem{vinyals2016matching}
O.~Vinyals, C.~Blundell, T.~Lillicrap, D.~Wierstra \emph{et~al.}, ``Matching
  networks for one shot learning,'' in \emph{NeurPIS}, 2016, pp. 3630--3638.

\bibitem{kirkpatrick2017overcoming}
J.~Kirkpatrick, R.~Pascanu, N.~Rabinowitz, J.~Veness, G.~Desjardins, A.~A.
  Rusu, K.~Milan, J.~Quan, T.~Ramalho, A.~Grabska-Barwinska \emph{et~al.},
  ``Overcoming catastrophic forgetting in neural networks,'' \emph{Proceedings
  of the national academy of sciences}, vol. 114, no.~13, pp. 3521--3526, 2017.

\bibitem{kaushik2021understanding}
P.~Kaushik, A.~Kortylewski, A.~Gain, and A.~Yuille, ``Understanding
  catastrophic forgetting and remembering in continual learning with optimal
  relevance mapping,'' in \emph{Fifth Workshop on Meta-Learning at the
  Conference on Neural Information Processing Systems}, 2021.

\bibitem{hou2019learning}
S.~Hou, X.~Pan, C.~C. Loy, Z.~Wang, and D.~Lin, ``Learning a unified classifier
  incrementally via rebalancing,'' in \emph{CVPR}, 2019, pp. 831--839.

\bibitem{chen2021incremental}
K.~Chen and C.-G. Lee, ``Incremental few-shot learning via vector quantization
  in deep embedded space,'' in \emph{ICLR}, 2021.

\bibitem{cheraghian2021semantic}
A.~Cheraghian, S.~Rahman, P.~Fang, S.~K. Roy, L.~Petersson, and M.~Harandi,
  ``Semantic-aware knowledge distillation for few-shot class-incremental
  learning,'' in \emph{CVPR}, 2021, pp. 2534--2543.

\bibitem{dong2021few}
S.~Dong, X.~Hong, X.~Tao, X.~Chang, X.~Wei, and Y.~Gong, ``Few-shot
  class-incremental learning via relation knowledge distillation,'' in
  \emph{AAAI}, 2021, pp. 1255--1263.

\bibitem{khrulkov2020hyperbolic}
V.~Khrulkov, L.~Mirvakhabova, E.~Ustinova, I.~Oseledets, and V.~Lempitsky,
  ``Hyperbolic image embeddings,'' in \emph{Proceedings of the IEEE/CVF
  Conference on Computer Vision and Pattern Recognition}, 2020, pp. 6418--6428.

\bibitem{ganea2018hyperbolic}
O.~Ganea, G.~B{\'e}cigneul, and T.~Hofmann, ``Hyperbolic neural networks,''
  \emph{Advances in neural information processing systems}, vol.~31, 2018.

\bibitem{nickel2017poincare}
M.~Nickel and D.~Kiela, ``Poincar{\'e} embeddings for learning hierarchical
  representations,'' \emph{Advances in neural information processing systems},
  vol.~30, 2017.

\bibitem{nickel2018learning}
------, ``Learning continuous hierarchies in the lorentz model of hyperbolic
  geometry,'' in \emph{International Conference on Machine Learning}.\hskip 1em
  plus 0.5em minus 0.4em\relax PMLR, 2018, pp. 3779--3788.

\bibitem{aly2019every}
R.~Aly, S.~Acharya, A.~Ossa, A.~K{\"o}hn, C.~Biemann, and A.~Panchenko, ``Every
  child should have parents: A taxonomy refinement algorithm based on
  hyperbolic term embeddings,'' in \emph{ACL (1)}, 2019.

\bibitem{tifrea2018poincare}
A.~Tifrea, G.~Becigneul, and O.-E. Ganea, ``Poincare glove: Hyperbolic word
  embeddings,'' in \emph{International Conference on Learning Representations},
  2018.

\bibitem{bachmann2020constant}
G.~Bachmann, G.~B{\'e}cigneul, and O.~Ganea, ``Constant curvature graph
  convolutional networks,'' in \emph{International Conference on Machine
  Learning}.\hskip 1em plus 0.5em minus 0.4em\relax PMLR, 2020, pp. 486--496.

\bibitem{liu2019hyperbolic}
Q.~Liu, M.~Nickel, and D.~Kiela, ``Hyperbolic graph neural networks,''
  \emph{Advances in Neural Information Processing Systems}, vol.~32, 2019.

\bibitem{liu2020hyperbolic}
S.~Liu, J.~Chen, L.~Pan, C.-W. Ngo, T.-S. Chua, and Y.-G. Jiang, ``Hyperbolic
  visual embedding learning for zero-shot recognition,'' in \emph{Proceedings
  of the IEEE/CVF conference on computer vision and pattern recognition}, 2020,
  pp. 9273--9281.

\bibitem{jain2014multi}
L.~P. Jain, W.~J. Scheirer, and T.~E. Boult, ``Multi-class open set recognition
  using probability of inclusion,'' in \emph{European Conference on Computer
  Vision}.\hskip 1em plus 0.5em minus 0.4em\relax Springer, 2014, pp. 393--409.

\bibitem{zhang2016sparse}
H.~Zhang and V.~M. Patel, ``Sparse representation-based open set recognition,''
  \emph{IEEE transactions on pattern analysis and machine intelligence},
  vol.~39, no.~8, pp. 1690--1696, 2016.

\bibitem{mendes2017nearest}
P.~R. Mendes~J{\'u}nior, R.~M. De~Souza, R.~d.~O. Werneck, B.~V. Stein, D.~V.
  Pazinato, W.~R. de~Almeida, O.~A. Penatti, R.~d.~S. Torres, and A.~Rocha,
  ``Nearest neighbors distance ratio open-set classifier,'' \emph{Machine
  Learning}, vol. 106, no.~3, pp. 359--386, 2017.

\bibitem{bendale2016towards}
A.~Bendale and T.~E. Boult, ``Towards open set deep networks,'' in
  \emph{Proceedings of the IEEE conference on computer vision and pattern
  recognition}, 2016, pp. 1563--1572.

\bibitem{yoshihashi2019classification}
R.~Yoshihashi, W.~Shao, R.~Kawakami, S.~You, M.~Iida, and T.~Naemura,
  ``Classification-reconstruction learning for open-set recognition,'' in
  \emph{Proceedings of the IEEE/CVF Conference on Computer Vision and Pattern
  Recognition}, 2019, pp. 4016--4025.

\bibitem{ge2017generative}
Z.~Ge, S.~Demyanov, Z.~Chen, and R.~Garnavi, ``Generative openmax for
  multi-class open set classification,'' in \emph{British Machine Vision
  Conference 2017}.\hskip 1em plus 0.5em minus 0.4em\relax British Machine
  Vision Association and Society for Pattern Recognition, 2017.

\bibitem{oza2019c2ae}
P.~Oza and V.~M. Patel, ``C2ae: Class conditioned auto-encoder for open-set
  recognition,'' in \emph{Proceedings of the IEEE/CVF Conference on Computer
  Vision and Pattern Recognition}, 2019, pp. 2307--2316.

\bibitem{perera2020generative}
P.~Perera, V.~I. Morariu, R.~Jain, V.~Manjunatha, C.~Wigington, V.~Ordonez, and
  V.~M. Patel, ``Generative-discriminative feature representations for open-set
  recognition,'' in \emph{Proceedings of the IEEE/CVF Conference on Computer
  Vision and Pattern Recognition}, 2020, pp. 11\,814--11\,823.

\bibitem{zhou2021learning}
D.-W. Zhou, H.-J. Ye, and D.-C. Zhan, ``Learning placeholders for open-set
  recognition,'' in \emph{Proceedings of the IEEE/CVF Conference on Computer
  Vision and Pattern Recognition}, 2021, pp. 4401--4410.

\bibitem{castro2018end}
F.~M. Castro, M.~J. Mar{\'\i}n-Jim{\'e}nez, N.~Guil, C.~Schmid, and K.~Alahari,
  ``End-to-end incremental learning,'' in \emph{ECCV}, 2018, pp. 233--248.

\bibitem{cheraghian2021synthesized}
A.~Cheraghian, S.~Rahman, S.~Ramasinghe, P.~Fang, C.~Simon, L.~Petersson, and
  M.~Harandi, ``Synthesized feature based few-shot class-incremental learning
  on a mixture of subspaces,'' in \emph{ICCV}, 2021, pp. 8661--8670.

\bibitem{krizhevsky2009learning}
A.~Krizhevsky, G.~Hinton \emph{et~al.}, ``Learning multiple layers of features
  from tiny images,'' 2009.

\bibitem{chaudhry2018efficient}
A.~Chaudhry, M.~Ranzato, M.~Rohrbach, and M.~Elhoseiny, ``Efficient lifelong
  learning with a-gem,'' in \emph{ICLR}, 2018.

\bibitem{he2016deep}
K.~He, X.~Zhang, S.~Ren, and J.~Sun, ``Deep residual learning for image
  recognition,'' in \emph{CVPR}, 2016, pp. 770--778.

\bibitem{robbins1951stochastic}
H.~Robbins and S.~Monro, ``A stochastic approximation method,'' \emph{The
  annals of mathematical statistics}, pp. 400--407, 1951.

\end{thebibliography}

\end{document}